%% file: Manuscript.tex
\documentclass[draft]{agujournal2019}

\usepackage{url}
\usepackage{lineno}
\usepackage[inline]{trackchanges}
\usepackage{soul}
\usepackage{lipsum}
\usepackage[utf8]{inputenc}
\usepackage{booktabs}
\usepackage{array} 
\usepackage{graphicx}
\usepackage{amsmath}
\usepackage[export]{adjustbox}
\usepackage{caption}
\usepackage{subcaption}
\usepackage{wrapfig}
\usepackage{tikz}
\usepackage{dsfont}

\newcommand{\indicator}{\mathds{1}}

\draftfalse

\journalname{JGR: Machine Learning and Computation}

\begin{document}

\title{HiPredict: Predicting the missing HiRISE RED4 spectral channel with machine learning}


\authors{A. Suphi\affil{1}, B. Moseley\affil{1}, G. S. Collins\affil{1}, V. T. Bickel\affil{2}, V. G. Rangarajan\affil{3}}

\affiliation{1}{Department of Earth Science and Engineering, Imperial College London, London, UK}
\affiliation{2}{Center for Space and Habitability, University of Bern, Bern, Switzerland}
\affiliation{3}{Institute of Space Research, German Aerospace Center (DLR), Berlin, Germany}

\correspondingauthor{Ardan Suphi}{ardan.suphi23@imperial.ic.ac.uk}


\begin{keypoints}
\item Since 2023, RED4, a crucial detector in the center of the HiRISE focal plane assembly, has failed intermittently.
\item We use machine learning to reconstruct RED4 from co-located spectral channels, outperforming the existing SYN4 product used in the HiRISE image processing pipeline.
\item Our reconstructed images demonstrate increased performance on downstream tasks including icy crater identification and spectral surface analysis. 
\end{keypoints}


\input{Sections/0.Abstract}

\input{Sections/1.Introduction}

\input{Sections/2.Background}

\input{Sections/3.Methodology}
\input{Sections/4.Results}
\input{Sections/5.Downstream}
\input{Sections/6.Discussion}


\section*{Data availability}

The HiPredict model weights \cite{suphi2026hipredictmodel} and the training and evaluation dataset \cite{suphi2026hipredictdataset} are available on Hugging Face under the MIT and CC BY 4.0 licences respectively, without access restrictions. The HiPredict software (v0.1.0), used for downloading, preprocessing, training, evaluation and inference, is preserved on Zenodo \cite{suphi2026hipredictcode} under the MIT licence and developed at https://github.com/ardan-S/HiPredict. The HiRISE observations used in this study are available from the NASA Planetary Data System \cite{HttpsHirisepdslplarizonaeduPDS}.

\section*{Conflict of Interest declaration}
The authors declare there are no conflicts of interest for this manuscript.

\acknowledgments
Many thanks to Shane, and the HiRISE team for their support and advice during this work. Especially Audrie, Chris and Rod for advising on the preprocessing pipeline and Ross for his support with the hiproc package. VGR thanks support from the Alexander von Humboldt Foundation. 


\newpage
\bibliography{references-final}

\end{document}


\title{Supporting Information for ``HiPredict: Predicting the missing HiRISE RED4 spectral channel with machine learning"}

\authors{A. Suphi\affil{1}, B. Moseley\affil{1}, G. S. Collins\affil{1}, V. T. Bickel\affil{2}, V. Rangarajan\affil{3}}

\affiliation{1}{Department of Earth Science and Engineering, Imperial College London, London, UK}
\affiliation{2}{Center for Space and Habitability, University of Bern, Bern, Switzerland}
\affiliation{3}{Institute of Space Research, German Aerospace Center (DLR), Berlin, Germany}

\begin{article}

%
%

\noindent\textbf{Contents of this file}
\begin{enumerate}
\item Examination of assumptions underlying the SYN4 model
\item Distribution of observations in the training and evaluation sets
\item HiPredict and SYN4 correlation of predicted pixels against their true value
\item Difference plots illustrating the error between HiPredict and the ground truth
\item Table of the observations and coordinates used in domain analysis
\end{enumerate}

\clearpage
\noindent\textbf{S1. Examination of assumptions underlying the SYN4 model}

\begin{center}
    \includegraphics[height=5cm]{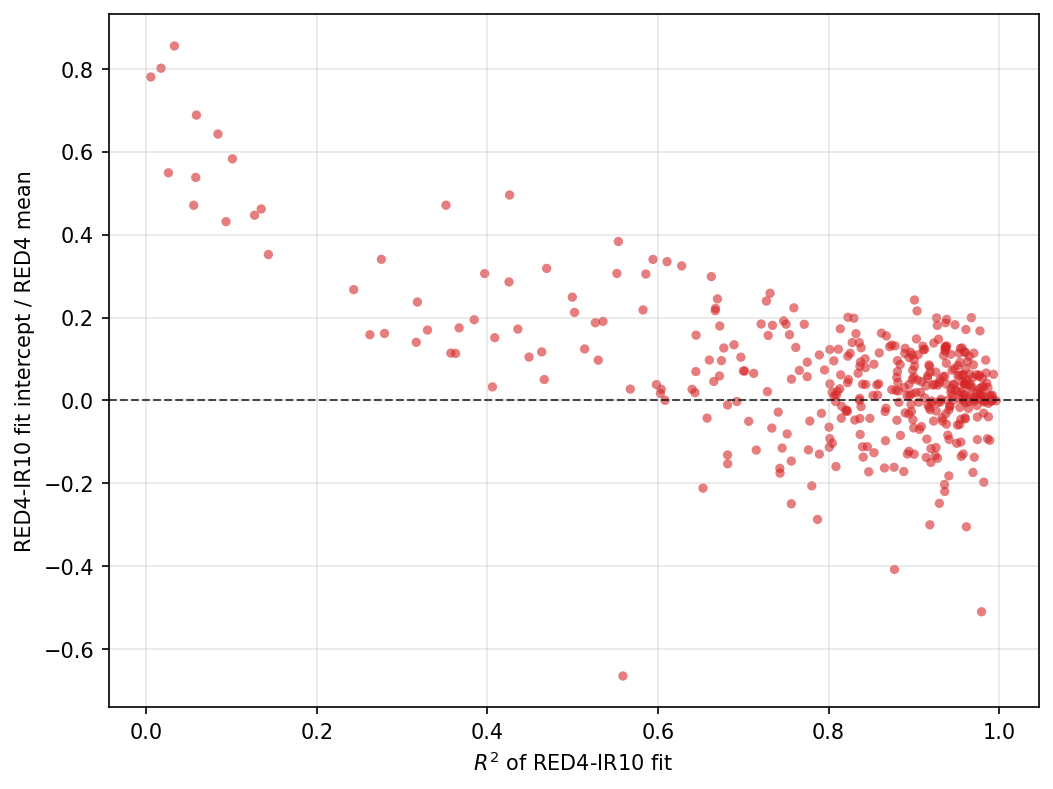}
    \hfill
    \includegraphics[height=5cm]{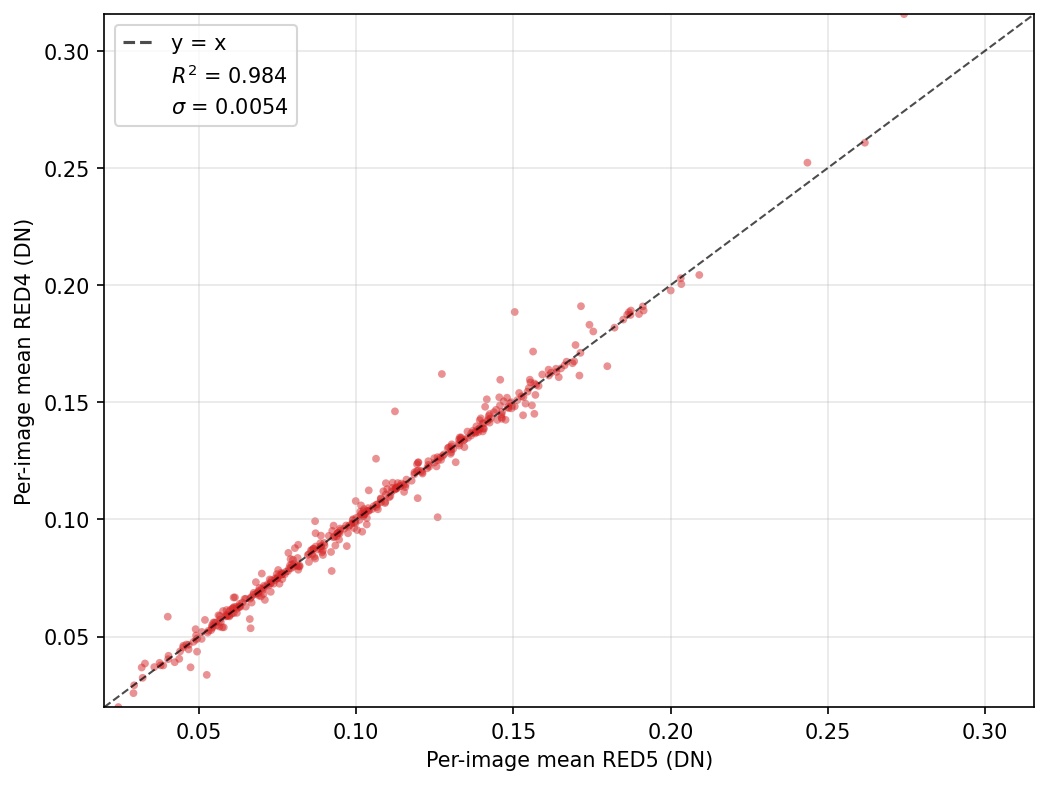}

    \vspace{1em}

    \textbf{Figure S1.} Visualisation of the SYN4 core assumptions.
\end{center}

    The SYN4 product rests on three assumptions: That the RED4-IR10 relationship is linear, that this linear relationship has no intercept and that the mean RED5 pixel value is a good proxy for that of RED4. The first two assumptions are examined in Figure A and the latter in Figure B.

    \textbf{(A)} A linear fit is applied to a selection of images and the $R^2$ value is plotted along the x-axis. Many points are highly correlated (closer to the right hand side) with a few clear exceptions where a linear fit is weak. The y-axis shows a normalised value of the intercept from this fit. Points which have a weak linear correlation also often have a large intercept. For many samples, both assumptions hold, but not always.
    
    \textbf{(B)} For the same selected images, the mean values of RED4 and RED5 are plotted against each other. This trend is clearly very strong, with fewer and weaker outliers.

\newpage
\noindent\textbf{S2. Distribution of observations in the training and evaluation sets}

\begin{center}
\includegraphics[width=0.48\textwidth]{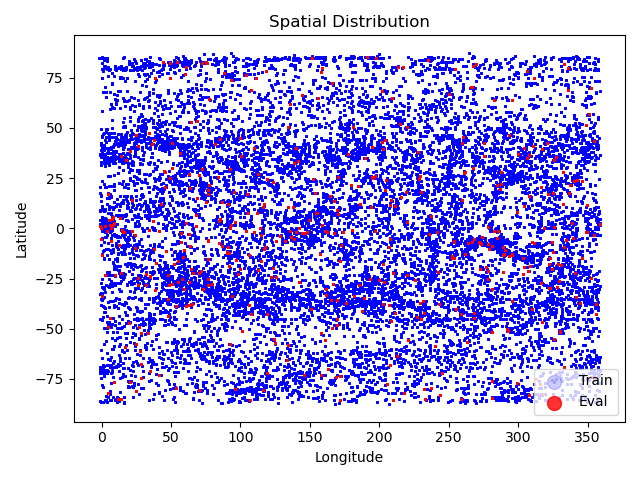}
\hfill
\includegraphics[width=0.48\textwidth]{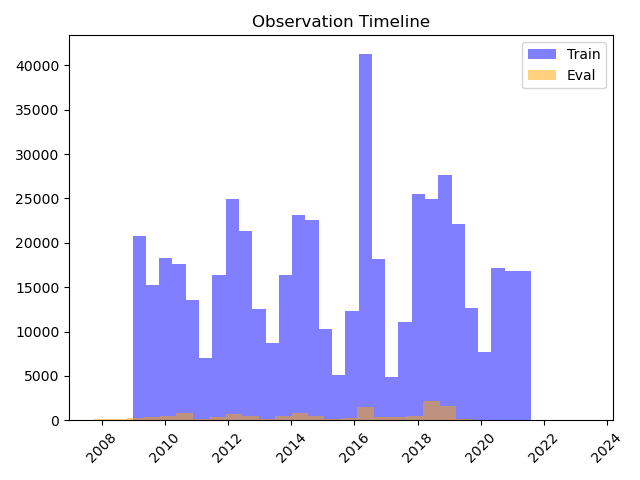}

\vspace{1em}

\noindent\textbf{Figure S2.} Spatial and temporal distribution of observations used in training and evaluation datasets.

\end{center}

\noindent\textbf{S3. HiPredict and SYN4 correlation of predicted pixels}

\begin{center}
\includegraphics[width=0.48\textwidth]{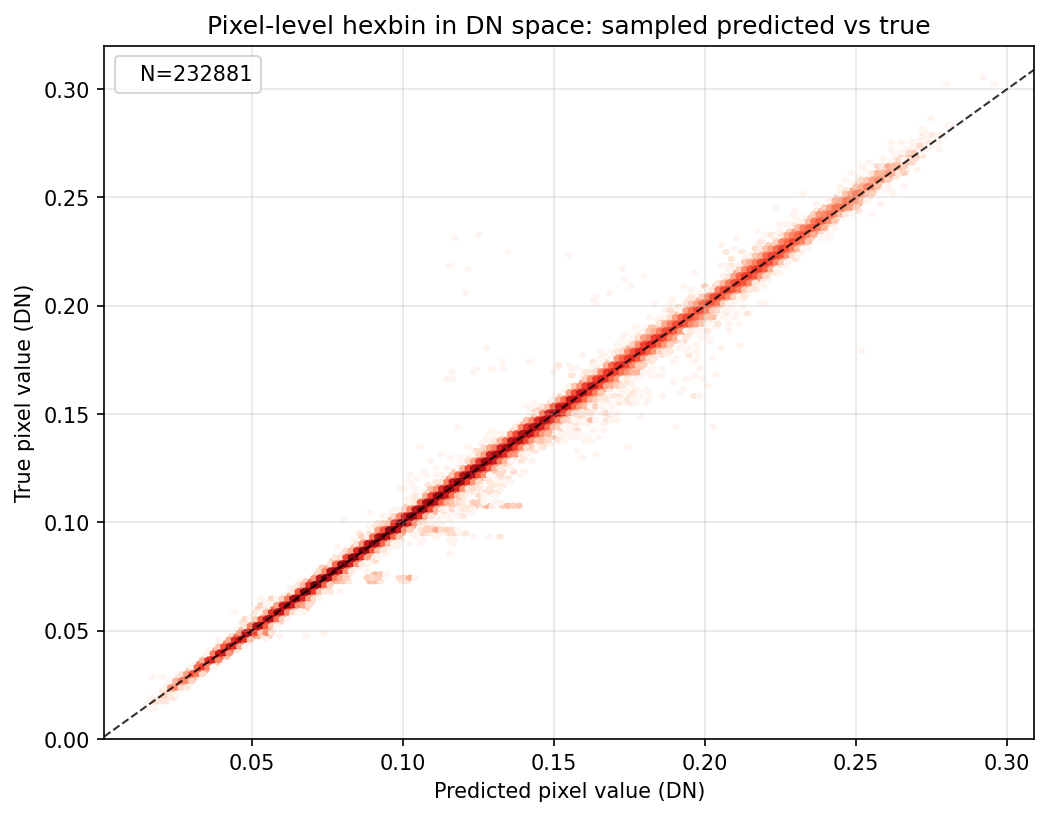}
\hfill
\includegraphics[width=0.48\textwidth]{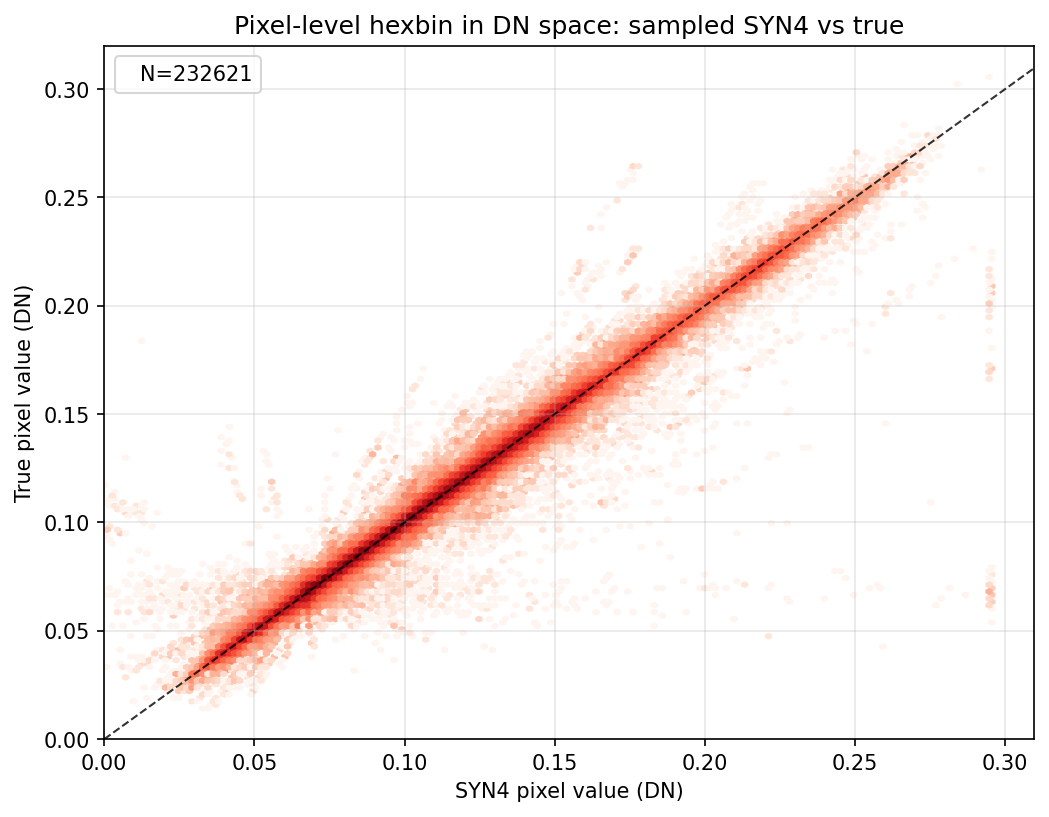}

\vspace{1em}

\noindent\textbf{Figure S3.} Correlation between predicted and true pixel values for HiPredict and SYN4.
\end{center}

HiPredict, shown on the left, has lower deviation and fewer outliers relative to SYN4, shown on the right. This illustrates that, in addition to improved overall performance, HiPredict produces fewer outliers and is more consistent than SYN4.

\clearpage
\noindent\textbf{S4. Diff plots illustrating the error between HiPredict and the ground truth}
\vspace{1em}
\begin{center}
\includegraphics[width=0.32\textwidth]{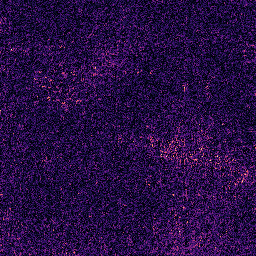}
\hfill
\includegraphics[width=0.32\textwidth]{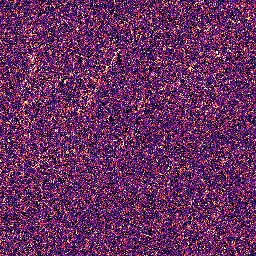}
\hfill
\includegraphics[width=0.32\textwidth]{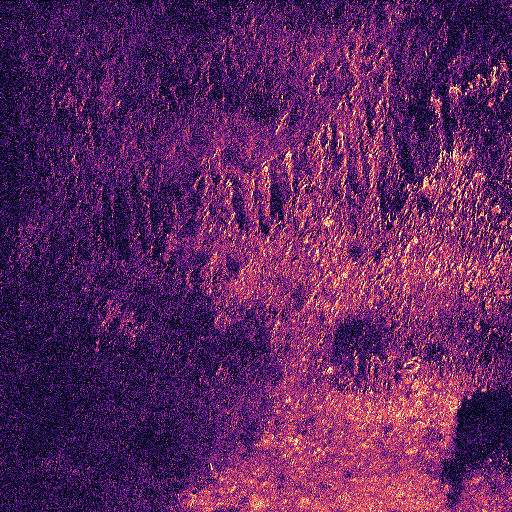}

\vspace{1.5ex}

\includegraphics[width=0.32\textwidth]{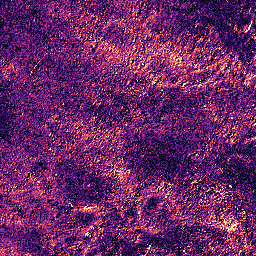}
\hfill
\includegraphics[width=0.32\textwidth]{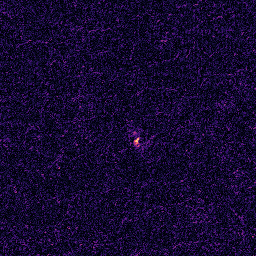}
\hfill
\includegraphics[width=0.32\textwidth]{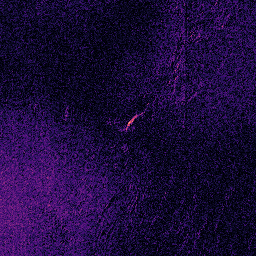}

\vspace{1em}

\noindent\textbf{Figure S4.} Difference plots showing the residual error between HiPredict and the ground truth for six observations. 
\end{center}
From these observations, much of the remaining residual structure resembles high-frequency `salt-and-pepper' style error, likely caused by instrument noise. While some local features can be discerned, for the most part this shows minimal remaining signal. Images are shown row-wise from top left to bottom right: ESP\_055232\_1595, ESP\_056705\_1440, ESP\_012940\_1655, ESP\_078908\_1735, ESP\_051070\_1285, and ESP\_077040\_1565.

\newpage
\noindent\textbf{S5. Table of the observations and coordinates used in domain analysis}

    \noindent\textbf{Table S1.} Observation IDs, image pixel coordiantes and feature classifications used in the domain analysis. The table is provided as separate supporting information file, \texttt{Table\_S1\_field\_observations.csv}

\noindent\textbf{S6. SHAP plot for metadata}

\begin{center}
    \includegraphics[height=15cm]{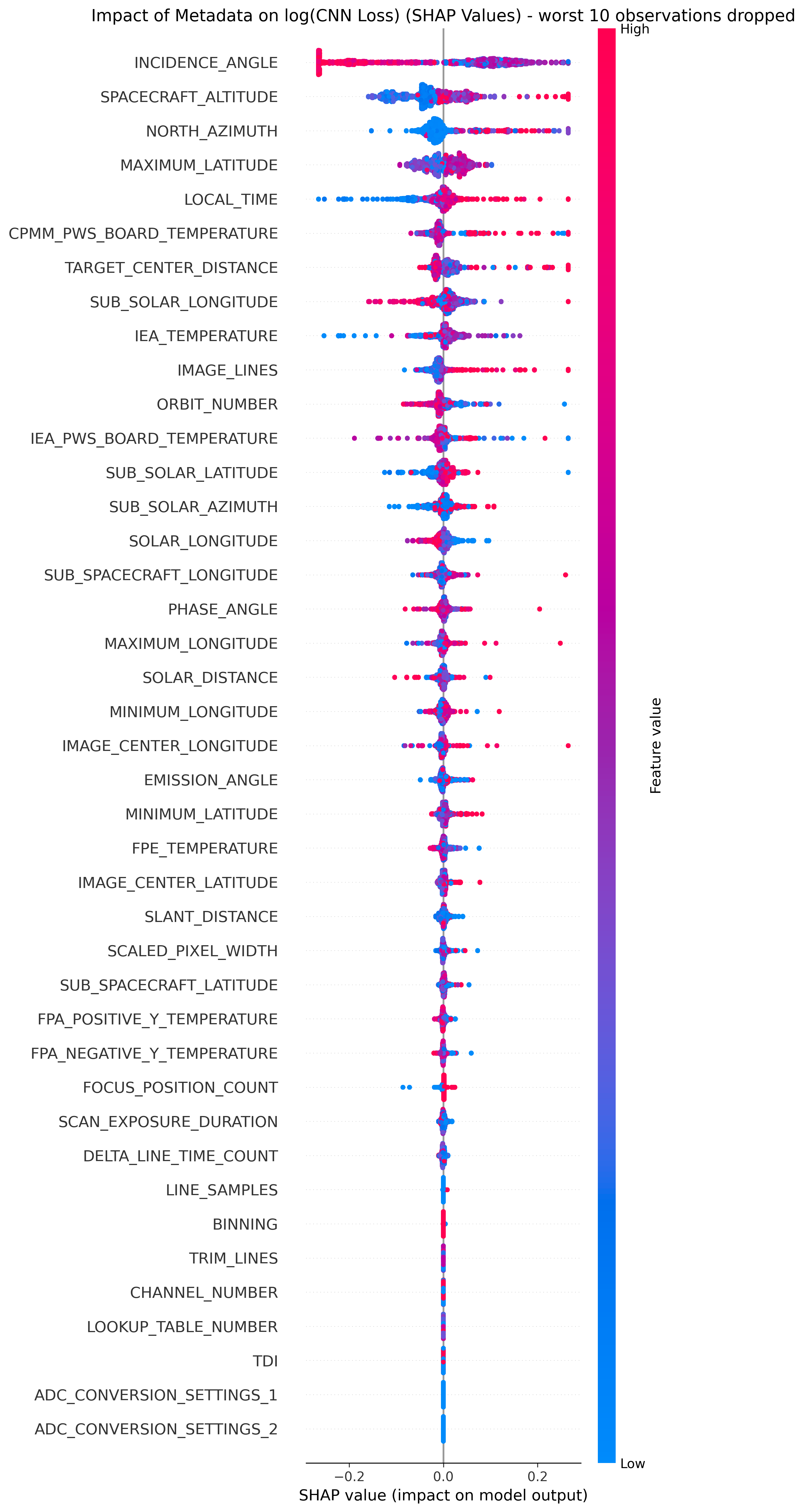}
\end{center}

SHAP plot showing the impact of each metadata value on model performance. Points to the left of center have a positive impact on model performance, points to the right a negative impact. Red points indicate a larger value for that metadata item and blue a lower value. Verticality indicates density of data points.  

\end{article}
\clearpage

%% file: Sections/0.Abstract.tex
\begin{abstract}

The High Resolution Imaging Science Experiment (HiRISE) onboard the Mars Reconnaissance Orbiter (MRO) acquires imagery of Mars at 25--30 cm/pixel, supporting studies of martian geology, climate history, and landing-site characterization. Its 14-detector focal plane assembly includes several overlapping detectors that enable the acquisition of three-channel colour-infrared images, which are important for identifying compositional variations and studying active surface processes. In 2023, a detector in the center of the focal plane assembly (RED4) began failing intermittently, creating a gap in acquired imagery and reducing the instrument's colour-imaging capability by half. To mitigate this loss, the HiRISE processing pipeline provides a synthetic red (SYN4) channel obtained by linearly scaling the co-located infrared (IR10) channel. However, this approach cannot capture observed non-linear relationships between the colour channels, limiting its accuracy and applicability. Here, we present HiPredict, a machine-learning framework that reconstructs RED4 from co-located IR10 and blue-green (BG12) imagery, neighbouring red-channel observations, and acquisition metadata. The model employs a U-Net architecture to learn non-linear spatiotemporal-spectral relationships between these inputs and is trained on 81,717 training samples, representing 13,625 observations. Evaluated on 23,348 test samples, representing 3,892 pre-failure observations, HiPredict reconstructs RED4 with a Normalised Mean Absolute Error of 1.1\%, substantially outperforming SYN4 (3.1\%). The reconstructed products improve colour image recovery and more closely represent features of interest such as icy craters, stratigraphy exposed along ridge crests and chloride deposits. We release HiPredict as an open software tool. More broadly, this work demonstrates the potential of machine learning to mitigate data loss as planetary missions age and naturally degrade. 

\end{abstract}

\section*{Plain Language Summary}

The High Resolution Imaging Science Experiment (HiRISE) camera onboard NASA's Mars Reconnaissance Orbiter has provided some of the highest-resolution images ever taken of Mars, helping scientists study the planet's surface, climate history, and potential landing sites for exploration missions. In 2023, part of a key detector began failing, creating a gap in the center of images and reducing HiRISE's ability to capture colour data. Colour images are particularly valuable because they help distinguish different surface materials and expose changes over time. In this study, we develop a machine-learning model called HiPredict that reconstructs the missing data using information from the camera's remaining colour channels and surrounding pixels. The model produces substantially more accurate reconstructions than the method currently used by the HiRISE processing pipeline and improves the quality of several scientific analyses, including identification of icy craters. This work demonstrates how machine learning can help to recover missing measurements and extend the scientific lifetime of spacecraft instruments.

%% file: Sections/1.Introduction.tex
\section{Introduction}

In 2001, the NASA Mars Exploration Program (MEP) established four major themes for Mars exploration: the search for past life, the study of martian climate, the investigation of surface and subsurface geology and geophysics, and preparation for human exploration \cite{MEPAGScientificGoalsObjectives2001}. These goals, reaffirmed as recently as 2025 \cite{MarsScienceGoals2025}, formed the core scientific justification for the Mars Reconnaissance Orbiter (MRO) which launched in August 2005 \cite{grafMarsReconnaissanceOrbiter2005}. A core capability required to achieve these goals is --- and remains ---  the capacity for high-resolution imaging of the martian surface from orbit. Consequently, the High Resolution Imaging Science Experiment (HiRISE) was selected as one of six instruments onboard MRO and began its operations in November 2006 \cite{mcewenMarsReconnaissanceOrbiters2007}. Since then, the HiRISE camera has taken over 103,000 images covering approximately 3.5\% of the martian surface and made crucial contributions to many scientific areas including geomorphologic surface characterization, landing-site assessment, rover/lander operations support, stereo and topography imagery and monitoring of active surface processes. These have resulted in over 2000 peer-reviewed science publications as of 2024 \cite{mcewenHighresolutionImagingScience2024}. As this imager nears twenty years of operations though, it has suffered from age-related degradation.

\begin{figure}
    \centering
    \begin{subfigure}[b]{0.32\textwidth}
        \centering
        \begin{tikzpicture}
            \node[anchor=south west, inner sep=0] (image) at (0,0)
                {\includegraphics[width=\textwidth, trim=0 2cm 0 2cm, clip]{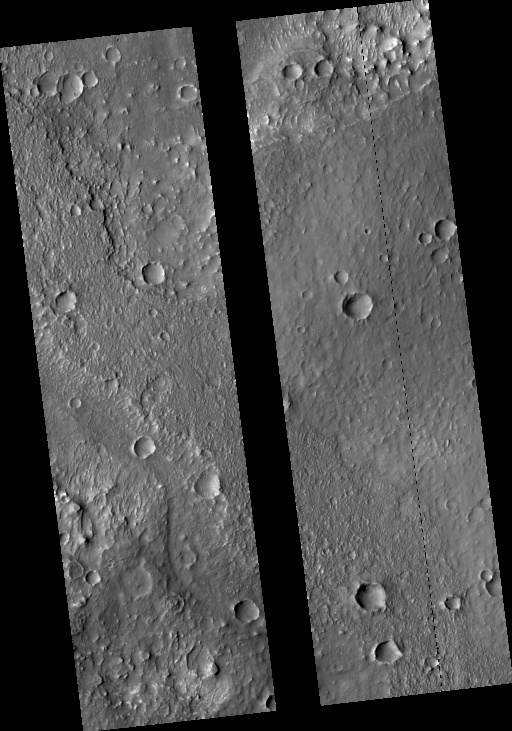}};
            \begin{scope}[x={(image.south east)}, y={(image.north west)}]
                \draw[white, line width=2pt] (0.05,0.06) -- (0.222,0.06);
                \node[white, anchor=south] at (0.136,0.075) {\small 1 km};
            \end{scope}
        \end{tikzpicture}
        \subcaption{ESP\_090807\_1460 ($-33.546^\circ, 350.844^\circ\text{E}$)}
        \label{fig:subfig:R4_gap_ex1}
    \end{subfigure}
    \hfill
    \begin{subfigure}[b]{0.32\textwidth}
        \centering
        \begin{tikzpicture}
            \node[anchor=south west, inner sep=0] (image) at (0,0)
                {\includegraphics[width=\textwidth, trim=0 14.5cm 0 2cm, clip]{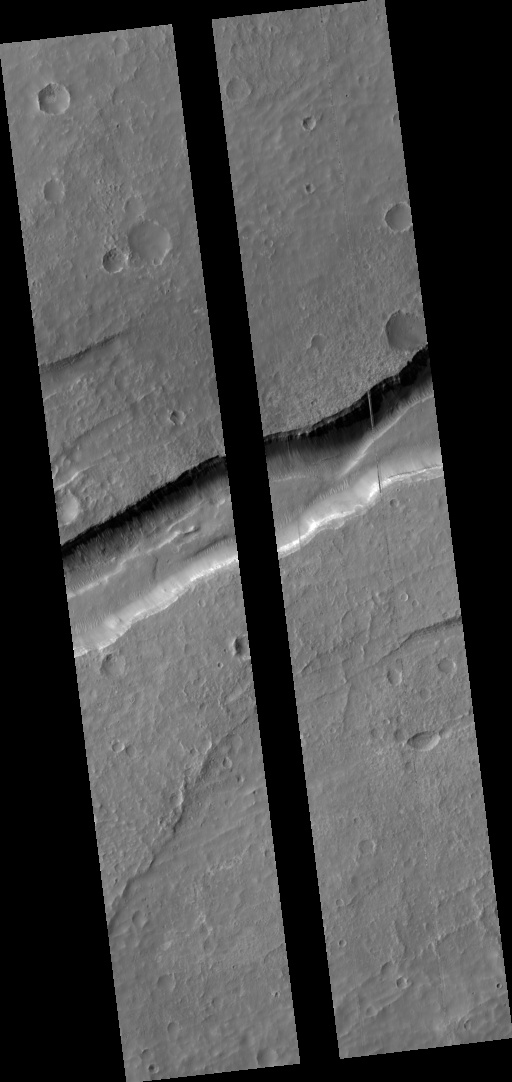}};
            \begin{scope}[x={(image.south east)}, y={(image.north west)}]
                \draw[white, line width=2pt] (0.05,0.06) -- (0.222,0.06);
                \node[white, anchor=south] at (0.136,0.075) {\small 1 km};
            \end{scope}
        \end{tikzpicture}
        \subcaption{ESP\_090826\_1430 ($-34.773^\circ, 192.493^\circ\text{E}$)}
        \label{fig:subfig:R4_gap_ex2}
    \end{subfigure}
    \hfill
    \begin{subfigure}[b]{0.32\textwidth}
        \centering
        \begin{tikzpicture}
            \node[anchor=south west, inner sep=0] (image) at (0,0)
                {\includegraphics[width=\textwidth, trim=0 15.3cm 0 2cm, clip]{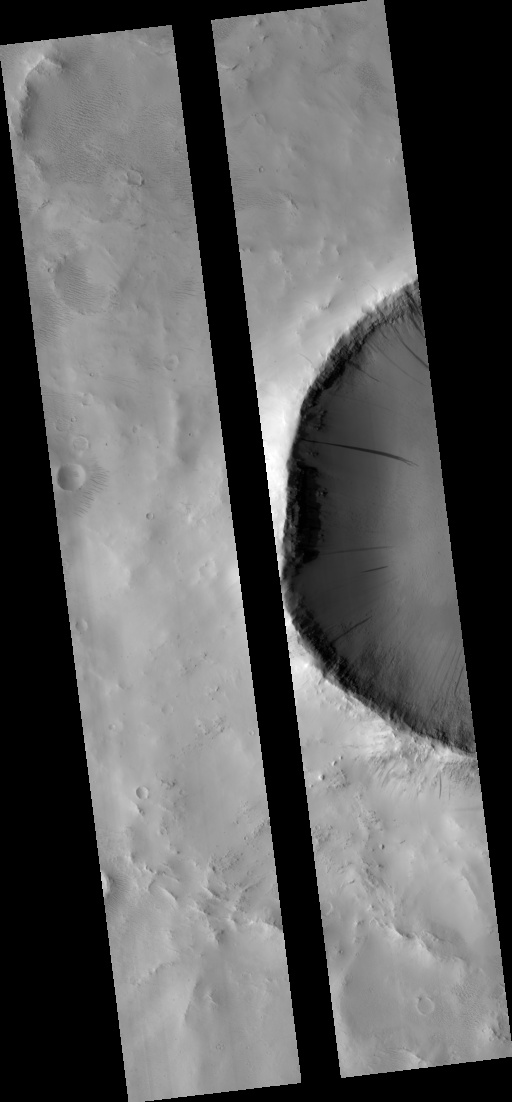}};
            \begin{scope}[x={(image.south east)}, y={(image.north west)}]
                \draw[white, line width=2pt] (0.05,0.06) -- (0.222,0.06);
                \node[white, anchor=south] at (0.136,0.075) {\small 1 km};
            \end{scope}
        \end{tikzpicture}
        \subcaption{ESP\_090845\_1840 ($4.101^\circ, 28.172^\circ\text{E}$)}
        \label{fig:subfig:R4_gap_ex3}
    \end{subfigure}
    \caption{Three recently-acquired HiRISE images (RED bandpass-only) showing the central gap that results from the RED4 failure. The gap induced by the RED9 failure is not directly visible.}
    \label{fig:red_gaps}
\end{figure}

\begin{figure}[htbp]
    \centering

    \begin{subfigure}[b]{\textwidth}
        \centering
        \hspace*{-1.5cm}    
        \begin{tikzpicture}
            \node[anchor=south west, inner sep=0] (image) at (0,0)
                {\includegraphics[width=0.77\textwidth, trim=0 0 0 0.5, clip]{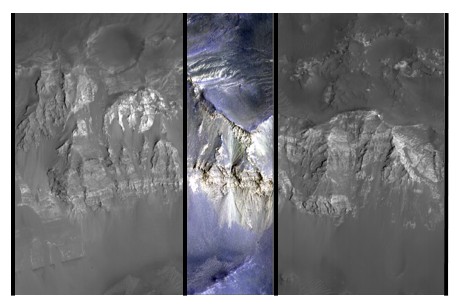}};
            \begin{scope}[x={(image.south east)}, y={(image.north west)}]
                \draw[black, line width=2pt] (0.05,0.06) -- (0.234,0.06);
                \node[black, anchor=south] at (0.142,0.075) {\small 1 km};
            \end{scope}
        \end{tikzpicture}
        \caption{An observation from before the RED9 and RED4 failures}
        \label{fig:subfig:schematic_example_full}
    \end{subfigure}

    \begin{subfigure}[b]{\textwidth}
        \centering
        \begin{tikzpicture}
            \node[anchor=south west, inner sep=0] (image) at (0,0)
                {\includegraphics[width=\textwidth, trim=0 0 0 0.5, clip]{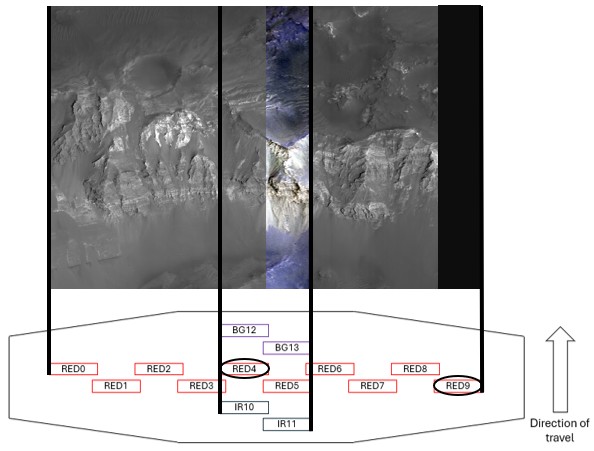}};
            \begin{scope}[x={(image.south east)}, y={(image.north west)}]
                \draw[black, line width=2pt] (0.1,0.38) -- (0.234,0.38);
                \node[black, anchor=south] at (0.167,0.393) {\small 1 km};
            \end{scope}
        \end{tikzpicture}
        \caption{The same observation, updated to represent the effect of the RED9 and RED4 failures (affected CCDs circled). The grey strip on the BG12/RED4/IR10 channel is the SYN4 product.}
        \label{fig:subfig:schematic_example_failed}
    \end{subfigure}

    \caption{A comparison showing the impact of the two CCD failures. As the BG12 and IR10 detectors overlap with RED4, this band can be predicted but as there is nothing which images in the RED9 spatial area, this data cannot be recovered. Observation ESP\_023753\_1660 ($-13.8236^\circ, 309.643^\circ\text{E}$).}
    \label{fig:Combined_focalplane_R4_gap_examples}
\end{figure}

The HiRISE focal plane contains 14 individual detectors. Ten detectors span the full swath in the red visible bandpass (550–-850 nm) with two infrared ($>$800 nm) and two blue-green filters ($<$600 nm) overlapping the two central red detectors, allowing for three-band colour-infrared imaging in this central swath. In August 2011, the leftmost red detector, RED9, failed completely, narrowing the image swath by $\sim$10\%. However, since July 2023, a central red detector, RED4, began failing intermittently \cite{akersMINDGAPHiRISE2025}, leaving a more problematic gap in the centre of the images (Figure \ref{fig:red_gaps}\subref{fig:subfig:R4_gap_ex1}--\subref{fig:subfig:R4_gap_ex3}). The RED4 failure is more problematic as it halves the area for taking colour-infrared images and creates a gap through the centre of the images, where most features of interest are located, illustrated in Figure \ref{fig:Combined_focalplane_R4_gap_examples}. In real terms, this gap corresponds to 500-600m on the surface. Features of interest on this order, such as small impact craters and outcrops, could be  missed entirely, or captured in only single-band, limiting their scientific return. Features tied to analyses of surface composition and/or (subsurface) ice and surface processes especially benefit from colour-infrared imagery. Additionally, the RED bandpasses are often acquired at a higher spatial resolution than the IR and BG bandpasses due to their superior Signal-to-Noise Ratio (SNR).

Efforts to mitigate the loss of RED4 have resulted in the publication of a synthetic product known as `SYN4'. This involves a linear scaling of the corresponding infrared image to match the mean pixel value of the neighbouring RED5 image. This has had some success \cite{akersMINDGAPHiRISE2025} however this approach cannot capture observed non-linear relationships between the colour channels, limiting its accuracy and applicability.

In this work, we use machine learning (ML) as an alternative way to predict the missing RED4 data. Specifically, we use a U-Net architecture with a custom brightness head to predict the gap from overlapping and surrounding images, as well as spacecraft, instrument, and environmental metadata. This allows us to capture non-linear relationships, train on spatial patches (rather than on a per-pixel basis), and incorporate environmental factors through the inclusion of metadata to generate more accurate predictions. We evaluate our model against several downstream tasks such as prediction accuracy of icy impacts, ridge spectra and chloride deposits. We evaluate these domains by applying our model to a subset of images containing these features to derive science domain-specific metrics to complement those from across the entire test set. We also provide a software tool to allow HiRISE users to reconstruct RED4 strips using their own inputs, and we are currently working with the HiRISE team to support the release of these products alongside future observations. By restoring colour information in the RED4/ IR10/ BG12 swath and improving the per-pixel accuracy of RED4 predictions, this work demonstrates the potential of virtual sensors and cross-bandpass extrapolation in planetary science using ML.

This paper is structured as follows: in Section \ref{sec:instrument_overview} we introduce the HiRISE instrument in more detail as well as the RED4 failure. In Section \ref{sec:HiPred_overview} we provide a brief overview of our model, HiPredict. Section \ref{sec:data_and_preproc} contains an overview of training data generation and the preprocessing pipeline to prepare HiRISE images for ML. Section \ref{sec:architecture_and_training} covers the model architecture in more detail as well as the training process. Section \ref{sec:other_models} covers the SYN4 model in more detail as well as other models attempted, Section \ref{sec:val_metrics} introduces evaluation metrics and Section \ref{method:downstream_eval} the downstream evaluation tasks. Sections \ref{sec:results} and \ref{sec:results_downstream} introduce the results and performance of the model with discussion in Section \ref{sec:discussion} and conclusions in Section \ref{sec:conclusion}.

\subsection{Related work}

Machine learning (ML) and computer vision (CV) methods are increasingly being used to extract information from planetary remote-sensing data. In planetary science, these approaches have shown particular promise for tasks such as geomorphic and compositional mapping, image denoising, super-resolution, stereo matching, and depth or topographic estimation \cite{nodjoumiDeepLandformsDeepLearning2023,moseleyLowlightImageEnhancement2020,taoMADNet20PixelScale2021}. These applications demonstrate the ability of data-driven models to learn spatial, textural, and contextual relationships that are difficult to capture using traditional methods. For HiRISE specifically, ML has previously been applied to problems such as surface-feature detection, terrain classification, image restoration, and automated analysis of large image archives \cite{wagstaffMarsImageContent2021}. 

The prediction of one spectral bandpass from another is a more specific and less widely studied problem, but it is closely related to several active areas of remote-sensing research. In planetary imaging, pan-sharpening and spectral-spatial fusion have been explored for instruments such as CaSSIS and CRISM, where information from bands or sensors with different spatial and spectral characteristics is combined to improve downstream analysis \cite{TULLO2026100121,wangDataFidelityorientedSpatialspectral2025}. Similar ideas have also been considered in heliophysics, where virtual or reconstructed channels have been used to mitigate limitations in solar-observing instruments \cite{jarolimDeepLearningFramework2025,szenicerDeepLearningVirtual2019a}. 

The closest related work to ours is \citeA{szenicerDeepLearningVirtual2019a}, who develop a `virtual replacement' for the failed MEGS-A spectrograph channel on NASA's Solar Dynamics Observatory (SDO). The authors trained a CNN-based machine learning model to infer these missing spectral measurements from contemporaneous images produced by SDO's Atmospheric Imaging Assembly. Both this work and the work in \citeA{szenicerDeepLearningVirtual2019a} can be framed as virtual instruments, exploiting complementary sensors to generate physically interpretable replacements for missing data. Aside from the domain and instruments, important differences include dimensionality and spatial accuracy.

%% file: Sections/2.Background.tex
\section{Background}
\label{sec:instrument_overview}

\subsection{The HiRISE camera}

The HiRISE camera operates in a TDI (Time Delay Imaging) pushbroom mode. It consists of 14 charged-coupled device detectors (CCDs), each with 128x2048 12 $\mu$m imaging pixels and adjacent CCDs overlapping by approximately 48 pixels\cite{delamereColorImagingMars2010}. This setup builds 2D images line-by-line as the spacecraft moves over Mars' surface. Images are acquired according to several pre-defined parameters that are commanded in advance, such as spatial binning which reduces the spatial resolution while increasing the SNR. The BG and IR images usually require higher binning to achieve the desired SNRs \cite{delamereColorImagingMars2010}. 

The structure of the focal plane can be seen in Figure \ref{fig:Combined_focalplane_R4_gap_examples}. The focal plane has ten broadband RED CCD detectors (labelled RED0-RED9), spanning the full swath. Four colour CCDs are present in the centre: IR10 and BG12 overlap with RED4 and IR11 and BG13 overlap with RED5 creating three-band colour coverage over the central fifth of the swath. 

\subsection{SYN4 image reconstruction}
\label{sec:syn4}

The HiRISE team’s operational solution to the RED4 failure is a synthetic RED4 product, known as `SYN4'. This is created by substituting IR10, adjusted in brightness to match RED5, into the missing RED4 location and is described in Equation \ref{eqn:syn4-equation} \cite{akersMINDGAPHiRISE2025}. This is the baseline model against which we evaluate this work.

\begin{equation}
\label{eqn:syn4-equation}
x_{\text{SYN4}}(i,j) = x_{\text{IR10}}(i,j) \cdot \frac{\mu\big(I_{\text{RED5}}\big)}{\mu\big(I_{\text{IR10}}\big)}
\end{equation}

Where $ x_I$ denotes a single pixel at location $(i, j)$ in image $I$ and $\mu_{\text{I}}$ denotes the mean pixel value across full image $I$. This model relies on three key assumptions:
\begin{enumerate}
    \item The IR10 and RED4 pixels are linearly correlated within a scene
    \item The relationship is pure gain with zero intercept
    \item That gain is recoverable from a neighbouring band (i.e. Terrain in a neighbouring RED CCD is a good proxy for terrain in RED4)
\end{enumerate}

In terms of visual, qualitative performance, SYN4 improves visual continuity and tone balance of the images, but discrepancies remain where there are strong RED-IR colour contrasts; importantly, RED-band RDRs provided via the PDS retain the gap, and standard IRB colour products currently do not incorporate SYN4 in the same way. While \citeA{akersMINDGAPHiRISE2025} describe the visual performance of SYN4, detailed evaluations and metrics for its performance over a large portion of HiRISE images are unavailable. SYN4 has been incorporated into selected PDS Extras and DTM production, but standard RED and COLOR RDR products retain the gap, and SYN4 is not intended as a quantitative replacement for RED4 \cite{akersMINDGAPHiRISE2025}. 

Published discussion of SYN4 remains limited, with little evidence so far of broad adoption in downstream science analyses. We note several structural drawbacks with the SYN4 product. Firstly, the underlying data still originates from the IR detector, meaning signal from all other wavelengths is systematically not acquired. We can therefore expect SYN4 to be spectrally representative of an IR product, not a RED one. This may be acceptable for RED mosaics and, for example, investigations of surface geomorphology, however it does not allow for spectrally accurate three-band colour images to be restored. Additionally, the data acquired by the IR and BG detectors are often binned, so the resulting SYN4 product is often at a coarser resolution than the other RED images.  Finally, as SYN4 is not a learned model, it is unaware of local or non-linear differences between the three spectral bands as well as the impact of environmental parameters like detector temperature.

%% file: Sections/3.Methodology.tex
\section{Methods}

\subsection{Overview of HiPredict}
\label{sec:HiPred_overview}

\begin{figure}
    \centering
    \begin{minipage}[c]{0.9\textwidth}
        \centering
        \includegraphics[width=\textwidth]{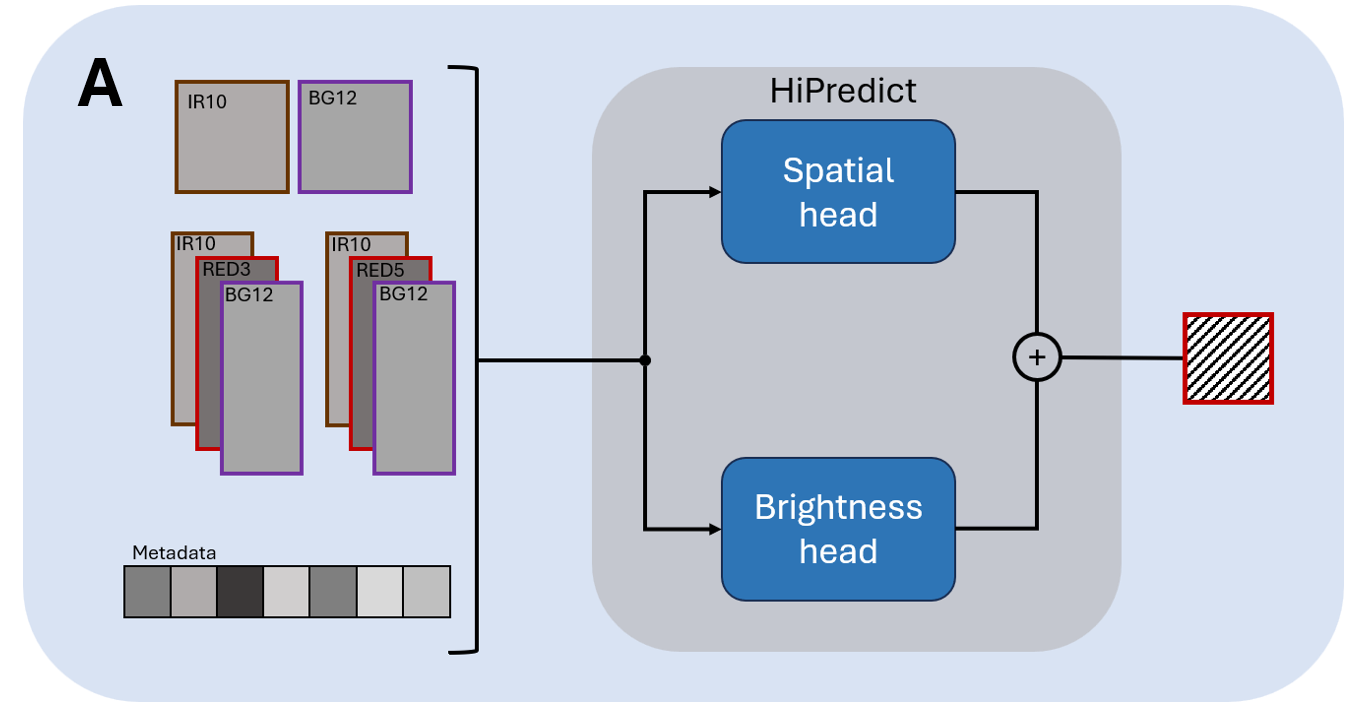}
    \end{minipage}

    \vspace{0.5cm}

    \begin{minipage}[c]{0.9\textwidth}
        \centering
        \includegraphics[width=\textwidth]{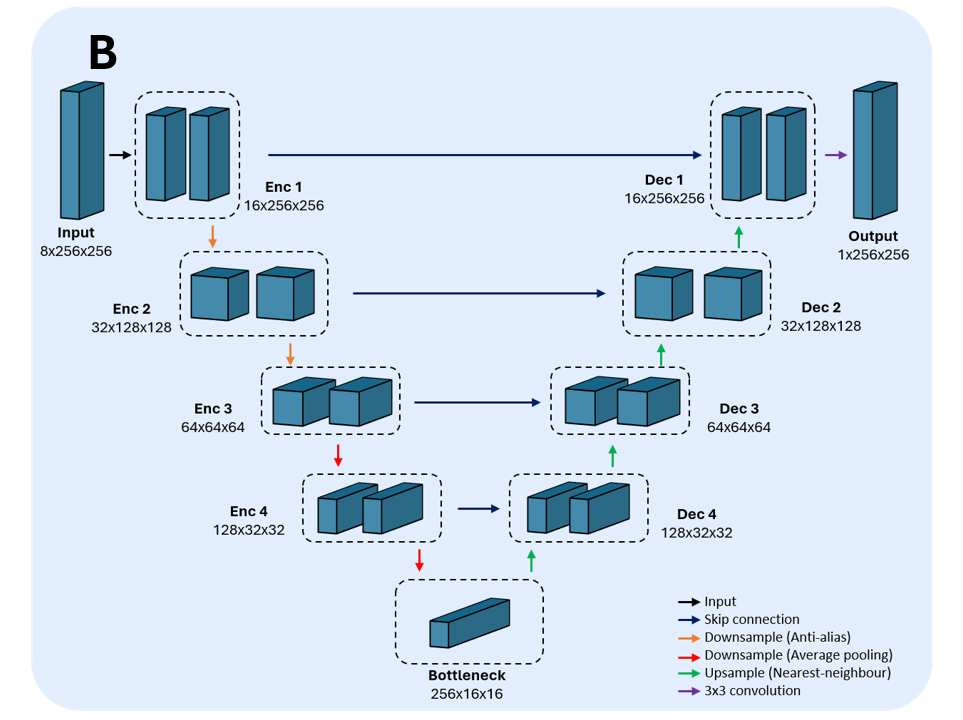}
    \end{minipage}

    \vspace{0.5cm}

    \begin{minipage}[c]{0.9\textwidth}
        \centering
        \includegraphics[width=\textwidth]{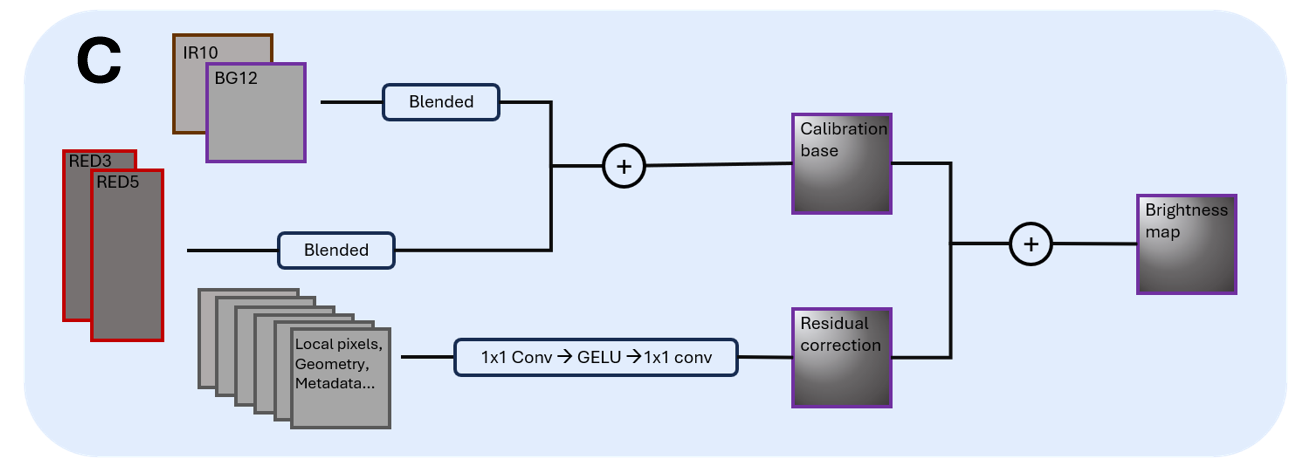}
    \end{minipage}

    \caption{Overview of HiPredict - (A) Overview of the HiPredict architecture showing the inputs and two model heads. The inputs, top to bottom, are the two band images from the target location, the overlap images from neighbouring CCDs, and metadata. (B) The spatial head, a U-Net based architecture. (C) The brightness head, a lightweight convolutional operator.}
    \label{fig:model_overview}
\end{figure}

We propose a machine learning based solution for predicting the RED4 output, called HiPredict. A high-level overview of our approach is shown in Figure \ref{fig:model_overview}. The goal of the model is to accurately reconstruct a patch of a RED4 HiRISE image from all available data. There are three inputs to this workflow:
\begin{enumerate}
    \item \textbf{Overlapping bands} -- Patches from the IR10 and BG12 detectors corresponding to the same spatial area as the predicted RED4 patch.
    \item \textbf{Neighbour images} -- We use the $\sim$48 pixel overlap on either side of the target RED4 strip to assemble narrow IR10/RED3/BG12 and IR10/RED5/BG12 arrays. 
    \item \textbf{Metadata} -- Single-value environment variables. 
\end{enumerate}
This data is passed through our model, and the output is a prediction for the missing RED4 patch. The model consists of two key components, a U-Net-based spatial head and a calibration-based brightness head. 

There are several benefits of a U-Net \cite{ronnebergerUNetConvolutionalNetworks2015} based approach. Firstly, the model is fully convolutional, which means it can be applied to images of any size at test-time. This allows for training to be done on smaller patches, which is crucial given HiRISE images can be tens of thousands of pixels long and are of variable sizes. Secondly, convolutional models have strong capacity to absorb spatial information, allowing this model to make pixel predictions with consideration to the pixel's surroundings. Finally, activation functions and hidden layers within the model allow for the modelling of the observed non-linear relationships between spectral bands.

The model is trained on 81,717 patches taken from real HiRISE observations from whilst RED4 was operational, and is evaluated on a hold-out set of 23,348 more recent patches. As well as testing on hold-out samples, we evaluate the model on a set of downstream tasks, based on true planetary science applications. We investigate HiPredict's performance at predicting icy impact craters, polygonal ridge spectra and chloride deposits. 

\subsection{Training Data Selection and Preprocessing}
\label{sec:data_and_preproc}

From the 95,500 observations publicly available as of November 2024, several criteria were applied to identify images suitable for training on. We first consider image metadata, some HiRISE images can be considered anomalous, such as those taken for calibration purposes or to capture uncommon events, like comet passes \cite{Wray2025HiRISE3IATLAS}. As these types of images are not the intended use-case for the model, we remove them by filtering out statistical outliers and keeping only standard detector encoding. We then apply eight acquisition/geometry conditions: Orbit number $\leq 100{,}000$; emission angle $\leq 15^\circ$; local time between 13:00 and 18:00; incidence angle between $0^\circ$ and $75^\circ$; binning mode is 1 or 2; TDI setting is 64 or 128; and mission phase is `PrimarySciencePhase` or `ExtendedSciencePhase'. Finally, we apply a completeness check to ensure all 5 desired CCDs are present. This filtering yielded 52,700 observations, from which a representative subset of 20,000 -- the maximum permitted by our computational constraints -- were selected for training. Of those 20,000 observations, 6 randomly selected patches of size 256x256 pixels were  saved, resulting in an estimated 120,000 training patches.

The images were obtained from the Planetary Data System (PDS) in Experimental Data Record (EDR) format \cite{mcewenPDSEDR2007}. This is the raw, unprocessed form of the images: following downlink from the spacecraft, only minor operations are applied (FELICS decompression, gap identification, pixel mirroring and metadata reconciliation \cite{SoftwareInterfaceSpecification2012}) to this product. The processing pipeline applied by the HiRISE team to create the Reduced Data Record (RDR) products, the final images, is described in \citeA{beckerHicalHiRISERadiometric2022} but we customise this workflow to generate training data. The pipeline can be considered in three stages: radiometric calibration, image stitching and geometric correction. For our training data, radiometric calibration is desirable as it minimises noise and maximises learnable signal. Geometric correction, however, was not implemented as interpolation and rotation steps could obscure underlying signal. The image stitching has no impact. 

The raw image files were first converted into cubefiles (.cub) and stitched to create CCD-level images. Radiometric calibration was applied (Isis3 command equivalent: hical \cite{ISIS9.0.0}), and then the RED images were stitched together (hiccdstitch). This then enables colour balancing of the IR and BG products (hicolorinit). Finally, jitter alignment is computed, registered and applied to ensure optimum spatial alignment of the RED CCDs (hijitreg and hislither). Following this, all images were converted into Python arrays, sampled into unique 256x256 patches and saved as training data. 

\subsection{HiPredict architecture and training}
\label{sec:architecture_and_training}

The HiPredict model is the final step in an iterative process to develop an accurate deep learning model to replace SYN4. An overview of its architecture can be seen in Figure \ref{fig:model_overview}. The core idea behind this architecture is to separate the spatial reconstruction from the radiometric and brightness predictions and enable these tasks through different parts of the model.

There are three categories of information streams being input into the model introduced in Section \ref{sec:HiPred_overview}. These are intended to provide different physical signals and inform the model in different ways. The \textit{band} data refers to the corresponding IR and BG images at the same location as the RED4 image which is being predicted. These carry high-frequency (HF) information across all spectral bands: details such as craters or boulders. The \textit{neighbour} data consists of two three-band stacks, one on either side of the target RED4 strip, making use of the $\sim$48 pixel overlap between CCDs. This contextual data can carry low-frequency (LF) information such as colour and brightness. They also contain a local view of the relationship between the three spectral bands, providing guidance to how the IR/BG bands can be converted into RED in the target location. The final data source is the metadata which provides instrument and environment level signal which can help the model calibrate noise and other factors.

The main branch is a residual U-Net \cite{ronnebergerUNetConvolutionalNetworks2015} whose role is to reconstruct the spatial detail in the missing RED4 image. The encoder-decoder structure with skip connections preserves fine spatial detail while allowing broader contextual reasoning. The absolute brightness/pixel values are handled separately by the brightness head, allowing the U-Net to focus purely on HF signal. In complex images, these resolution changes can introduce aliasing, checkerboard structure or block-like artefacts in image reconstruction tasks. HiPredict uses deliberately conservative down- and upsampling choices to reduce high-frequency folding into coarser feature maps. This is more common in tasks aiming to preserve and reconstruct continuous surface texture. The final step of the main branch computes a local spatial mean of the raw output and subtracts it from the pixel values to produce a locally centred detail field. This step explicitly separates HF detail from LF, radiometric and brightness factors. To ensure no seam or edge artefacts when stitching together large HiRISE images, the local mean is computed with a sliding spatial average window so is not based on global values and is translationally invariant. 

HiPredict estimates the LF brightness field with a separate brightness head. This branch restores the brightness level after the U-Net output has been centered. This head starts with the local IR10/BG12 averages, the side-detector calibration context, detector geometry and metadata. These inputs constitute an interpretable first-order estimate of the expected brightness through a simple calibration baseline: the starting value is the midpoint of local IR10 and BG12 brightness, adjusted by the blended neighbour calibration map. This first estimate is then calibrated by a learned delta which adjusts the baseline given other context. This residual correction to the initial estimate accounts for remaining nonlinear relationships, not captured by the simple calibration rule. 

The two architectural outputs are then summed to form the final RED4 prediction. The locally centred U-Net provides the spatial detail field and the brightness head provides the local brightness field.

The final model uses approximately 2.86 million trainable parameters, nine metadata variables and outputs a single RED4 channel image. The base number of channels is 16 with a U-Net model depth of 4 to a bottleneck dimension of 256. No dropout was used. The loss function was a weighted sum of Mean Absolute Error, also known as L1 (0.35), the L1 value of per-image means (0.3), Gradient-Difference loss (0.15), local L1 (0.1) and low frequency L1 (0.1). The AdamW optimiser was used with default parameters with a weight decay of $5e-5$ and a constant learning rate of $1.5e-3$. The brightness head was trained with twice the learning rate and the brightness head and global bias were warm-started for 2 epochs before the full network was unfrozen. The batch size was 64. Normalisation was done using a single global affine: $(x - center) / scale$, applied identically to inputs, labels and inversely to outputs. The center was 0.06787 and scale was 0.03831, derived from over $9.9\times10^{11}$ pixels across 13,625 observations. 

\subsection{Baseline models}
\label{sec:other_models}

Before settling on the final architecture, we developed a series of baselines against which we compare our final model. 

\begin{enumerate}
    \item \textbf{Benchmark model: SYN4} -
    As discussed in Section \ref{sec:syn4}, we compare to SYN4. We apply our own implementation of this algorithm to the denormalised images to recreate this product. 

    \item \textbf{Linear model} -
    The first model we attempted was direct linear mapping. This was considered the simplest `learned' model and the closest learned relative to SYN4. The linear model combines IR10 with a  reference from RED5. The model uses a learned 1x1 convolution over the two input channels to try and learn a direct local relationship between IR10 intensity and neighbouring brightness. This model contains three trainable parameters and is basic enough to be represented mathematically, as seen in Equation \ref{eqn:linear}.
    \begin{equation}
        \label{eqn:linear}
         RED4_{pred} = w_{ir}* IR10 + w_{ref}*RED_{ref} + b
    \end{equation}
    Where $w_{ir}$, $w_{ref}$ and $b$ are learnable parameters. This is a minimal brightness transfer baseline and not spatially expressive. By contrast, SYN4 has no trainable parameters but computes a new scale factor for each image. This model tests whether much of the RED4 reconstruction problem can be solved by simple radiometric scaling.

    \item \textbf{Linear model with metadata} -
    One of the limitations of linear model is that instrument and environment conditions could obscure a valid linear relationship. As such, a similar model was trained but including nine metadata values. This new approach is seen in Equation \ref{eqn:linear_w_meta}
    \begin{equation}
        \label{eqn:linear_w_meta}
         RED4_{pred} = w_{ir}* IR10 + w_{ref}*RED_{ref}+ w_{1}*meta_{1}+...+ w_{9}*meta_{9} + b
    \end{equation}
    With the capacity to correct for basic environmental changes and instrument noise, a linear radiometric relationship would be well modelled.

    \item \textbf{Basic U-Net} -
    The U-Net architecture is a dependable and reliable model and consequently, was a good out-of-the-box baseline for a nonlinear, spatially expressive model. It uses an encoder-decoder architecture with skip connections to reconstruct spatial structure. 
    It resembles the HiPredict spatial head but does not contain a separate brightness mechanism. It builds on the previous  models by introducing nonlinearity through hidden layers and activation functions. The input is 11 channels across the band and neighbour inputs, not including the metadata.

    \item \textbf{Custom U-Net} -
    This model introduces separate logic for brightness calibration. The U-Net output is locally centred by subtracting a sliding spatial mean, and a dedicated 1x1 mean convolutional pass then predicts DC level from locally-averaged input channel statistics. 
    The spatial improvements of this model over the basic U-Net are the introduction of anti-alias downsampling and nearest-neighbour upsampling with box-smoothing replacing transposed connections. This model is also slightly deeper with 4 stages instead of 3, and metadata is now consumed.
    This differs from the final model by not being fully convolutional (using batch normalisation which would cause seams on large inference tasks) and the HiPredict brightness head is a multi-layer perceptron, rather than a single convolutional layer.
\end{enumerate}

\subsection{Validation metrics}
\label{sec:val_metrics}

The model was assessed using a small set of complementary pixel-level, structural, and directional metrics. Lower values indicate better performance unless stated otherwise. These metrics were either applied at train-time, to evaluate model training, or at test-time, applied to a held-out subset of images. 

The metric equations are defined as follows: let $\{I^{(i)}\}_{i=1}^{N}$ denote the set of reference images and $\{\hat{I}^{(i)}\}_{i=1}^{N}$ the corresponding predictions.
Each image is defined on a discrete 2D grid of size $H \times W$, with pixel indices $x \in \{1,\dots,H\}$, $y \in \{1,\dots,W\}$.
We denote pixel intensities as $I^{(i)}_{xy}$ and $\hat{I}^{(i)}_{xy}$.

\begin{itemize}
    \item \textbf{Mean Absolute Error (MAE) and NMAE} measure the average absolute pixel error between the prediction and reference image across all images and spatial locations. MAE (Equation \ref{eqn:mae}) is used directly during training, while NMAE (Equation \ref{eqn:nmae}) normalises this error by the image dynamic range, allowing performance to be compared across scenes with different radiometric scales. 

    \begin{equation} 
    \mathrm{MAE} = \frac{1}{NHW} \sum_{i=1}^{N} \sum_{x=1}^{H} \sum_{y=1}^{W}
    \left| \hat{I}^{(i)}_{xy} - I^{(i)}_{xy} \right|
     \label{eqn:mae}
     \end{equation}

    \begin{equation}
    \mathrm{NMAE} = \frac{1}{N} \sum_{i=1}^{N}
    \frac{\sum_{x=1}^{H} \sum_{y=1}^{W} \left| \hat{I}^{(i)}_{xy} - I^{(i)}_{xy} \right|}
    {HW\,(I^{(i)}_{\max} - I^{(i)}_{\min})}
    \label{eqn:nmae}
    \end{equation}

    \item \textbf{Mean, local, and low-frequency MAE} are training-time variants of MAE that separate different sources of error. Mean MAE targets the recovery of the overall patch brightness; local MAE measures the reproduction of mean-subtracted spatial detail; and low-frequency MAE emphasises coarse shading and brightness-gradient structure. These terms encourage the model to match both the radiometric level and spatial structure of the target image.

    \begin{equation}
    \mathrm{MAE}_k = \frac{1}{NHW} \sum_{i=1}^{N} \sum_{x=1}^{H} \sum_{y=1}^{W}
    \left| \phi_k(\hat{I}^{(i)})_{xy} - \phi_k(I^{(i)})_{xy} \right|
    \label{eqn:multi-mae}
    \end{equation}
    
    where \(\phi_k(\cdot)\) denotes a scale-specific filtering operator (mean, local, or low-frequency), separating errors across spatial scales.

    \item \textbf{SSIM} measures structural similarity between the predicted and reference images, accounting for luminance, contrast, and local structure. It is used as an evaluation metric because it provides a bounded and interpretable measure of image fidelity that complements pixel-wise error metrics. Higher values indicate better agreement, with 1 corresponding to perfect structural similarity.

    \begin{equation}
    \mathrm{SSIM} = \frac{1}{N} \sum_{i=1}^{N} \mathrm{SSIM}(\hat{I}^{(i)}, I^{(i)})
    \end{equation}
    
    where:
    
    \[
    \mathrm{SSIM}(\hat{I}^{(i)}, I^{(i)}) =
    \frac{(2\mu_{\hat{I}^{(i)}}\mu_{I^{(i)}} + c_1)(2\sigma_{\hat{I}^{(i)} I^{(i)}} + c_2)}
    {(\mu_{\hat{I}^{(i)}}^2 + \mu_{I^{(i)}}^2 + c_1)(\sigma_{\hat{I}^{(i)}}^2 + \sigma_{I^{(i)}}^2 + c_2)}
    \]
    with \(\mu\) denoting mean intensity, \(\sigma^2\) variance, and \(\sigma_{\hat{I} I}\) covariance over local windows. SSIM measures perceptual similarity in structure, contrast, and luminance.

    \item \textbf{Percentage Within Tolerance (PWT 1\%)} reports the fraction of pixels whose predicted value lies within 1\% of the reference value. This provides an intuitive measure of how much of the reconstructed image is locally accurate, rather than summarising only the average error. Higher values indicate better performance.

    \begin{equation}
    \mathrm{PWT}_{1\%} = \frac{1}{NHW} \sum_{i=1}^{N} \sum_{x=1}^{H} \sum_{y=1}^{W}
    \indicator\!\left( \frac{\left| \hat{I}^{(i)}_{xy} - I^{(i)}_{xy} \right|}{I^{(i)}_{xy}} < 0.01 \right)
    \end{equation}

    where \(\indicator(\cdot)\) is the indicator function: A function which returns 1 or 0 depending on whether the statement inside is true. 
    
    \item \textbf{Gradient Difference Loss (GDL)} compares horizontal and vertical image gradients between the prediction and reference image. It is used only during training to reduce the over-smoothing commonly associated with MAE-based objectives and to encourage sharper, better-located edges. 

    \begin{equation}
    \makebox[0pt][c]{%
    \hspace*{-0.14\textwidth}%
    $\displaystyle
        \mathrm{GDL} = \frac{1}{N} \sum_{i=1}^{N} \sum_{x=1}^{H} \sum_{y=1}^{W}
        \left(
        \left| \hat{I}^{(i)}_{x+1,y} - \hat{I}^{(i)}_{xy}
        - (I^{(i)}_{x+1,y} - I^{(i)}_{xy}) \right|
        +
        \left| \hat{I}^{(i)}_{x,y+1} - \hat{I}^{(i)}_{xy}
        - (I^{(i)}_{x,y+1} - I^{(i)}_{xy}) \right|
        \right)
    $%
    }
    \end{equation}
        
\end{itemize}

\subsection{Downstream evaluation}
\label{method:downstream_eval}

We evaluate the model at three levels of depth and detail. The shallowest and broadest evaluation is the evaluation on the test set using metrics described in the previous Section. Once trained and evaluated on the test set, we move one step deeper by assessing the model's ability to predict common planetary and Mars science domains, using a subset of the test set. 

 We select three science themes and examine model performance for a selection of images in each of these domains. The themes are: icy impacts, spectral signatures exposed along polygonal ridge outcrops and chloride deposits. We compute the metrics used for the wider test set for each of these subsets, and visually inspect each of these outputs. We evaluate in this way because visual inspection of small-scale (colour) features is a core use of HiRISE imagery among planetary scientists and these domains represent a broad spectrum. Visual evaluation is important as the three selected themes are extremely rare in the training dataset, if not completely absent, so this is not only a measure of model performance in these narrow domains, but also a test of the model's generalisability and robustness. 

The third level of evaluation is the deepest and therefore the narrowest. With this detail, we investigate model performance only on icy craters, i.e., small-, (deca)meter-scale craters that form in response to a meteoroid impact and that expose ice from the subsurface. We choose this particular application because it requires both high spatial resolution (to identify the usually small lumps of ice) and spectral accuracy (to correctly identify water ice). For each evaluation image, we select an icy and non-icy pixel and evaluate the performance of SYN4 and HiPredict relative to the true value. Here, the true value is the spectra and band ratios derived using the original RED4 band. The procedure applied to identify the respective pixels and to compute the spectra and band ratios strictly follows \citeA{RANGARAJAN2024115849}. This assessment tells us not only the accuracy of reconstruction, but that underlying spectral relationships, such as the ability to classify icy pixels, has been preserved during inference. 

%% file: Sections/4.Results.tex
\section{Results}
\label{sec:results}

In this section we first show qualitative and quantitative analysis of our training data curation process. Then we report model training accuracy and benchmark performance, assessing model robustness across different operating conditions. Finally we present the performance of our model across the downstream tasks, and the integration of our model within production HiRISE processing pipelines.

\subsection{Training dataset curation and analysis}

Of the complete HiRISE catalog, 20,000 observations were identified to be included in the training dataset. Of these, 535 were unsuccessful during radiometric preprocessing  and 19,465 were successfully processed. From each of these observations, six non-overlapping patches of 256x256 pixels were selected and saved, yielding a total training data volume of 116,790 patches, divided into train (70\%), test (20\%) and validation (10\%) sets. These were split by age with the oldest, and therefore highest quality, present in the train set and the newest, containing the most noise, in the test set. The most common processing failures were HiCal (radiometric calibration) and HiJitReg (jitter registration) failures. The overall training dataset size was 202 GB. 

To develop the best model for the most common use cases, only the four most common binning and TDI combinations were used. These combinations were binning values of 1x1 and 2x2 and TDI values of 64 and 128. These combinations account for 92\% of all published HiRISE observations. The breakdown of the training data composition compared to the true dataset can be seen in Table \ref{tab:tdi_bin}.

\begin{table}[ht]
\centering
\begin{tabular}{lcccccc}
\toprule
& \multicolumn{3}{c}{\textbf{TDI (\%)}} 
& \multicolumn{3}{c}{\textbf{Binning (\%)}} \\
\cmidrule(lr){2-4}\cmidrule(lr){5-7}
& \textbf{128} & \textbf{64} & \textbf{Other}
& \textbf{1} & \textbf{2} & \textbf{Other} \\
\midrule
Train & 75.2 & 24.8 & 0.0 & 34.0 & 66.0 & 0.0 \\
True  & 70.2 & 24.4 & 5.3 & 30.6 & 65.8 & 3.6 \\
\bottomrule
\end{tabular}
\caption{TDI and binning distributions of the training and full HiRISE datasets.}
\label{tab:tdi_bin}
\end{table}

When examining the preprocessed images (our training data), we can investigate the SYN4 assumptions noted in Section \ref{sec:syn4}. We generally see a strong positive correlation between the per-image RED5 and RED4 mean values. Analysis on a subset of the training data yielded an $R^2$ value of 0.984 and standard deviation ($\sigma$) of 0.0054. There are notable outliers in this data but the relationship is generally strong. 

The assumptions relating to the IR10-RED4 correlation, however, are weaker. The linear characterisation of this relationship is weakest with mean $R^2$ between IR10 and RED4 as 0.81. We find a large $\sigma$ (0.21) and a long low tail (only 24\% of images clear $R^2$ $\geq$ 0.95, 1.8\% $\geq$ 0.99). This implies linearity may hold for the typical image but breaks down significantly for a meaningful minority, likely contributing to the high variance in SYN4 performance. Regarding the zero-intercept assumption we see the data is not centred at zero. The mean offset is +5.4\%, median +3.9\% of RED4 mean and a high $\sigma$ (0.16), indicating a systematic positive offset, not just symmetric scatter. 

These data reflects observations of SYN4 performance where we find some scenes perform exceptionally well but more complex scenes lead to a high error. These observations are visualised in Supplemental Information S3.

\subsection{Quantitative validation accuracy and baseline comparison}
\label{sec:val_accuracy_and_baseline_comparison}

We predict missing RED4 strips with a Normalised Mean Absolute Error (NMAE) of 1.11\%. As there is a wide range of pixel values within HiRISE images due to TDI and other commanded parameters, we present this error as a percentage of the image-level dynamic range. The HiPredict NMAE is 2.8 times smaller than the SYN4 value calculated on the same dataset.

\begin{figure}[h]
    \centering

    \begin{tikzpicture}[baseline=(image.center)]
        \node[anchor=south west, inner sep=0] (image) at (0,0)
            {\includegraphics[height=3.75cm]{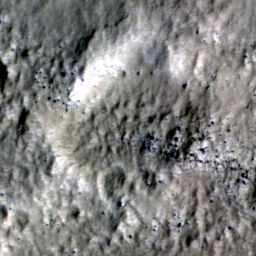}};
        \begin{scope}[x={(image.south east)}, y={(image.north west)}]
            \draw[white, line width=2pt] (0.05,0.06) -- (0.2069,0.06);
            \node[white, anchor=south] at (0.12845,0.075) {\small 20 m};
        \end{scope}
    \end{tikzpicture}
    \hfill
    \begin{tikzpicture}[baseline=(image.center)]
        \node[anchor=south west, inner sep=0] (image) at (0,0)
            {\includegraphics[height=3.75cm]{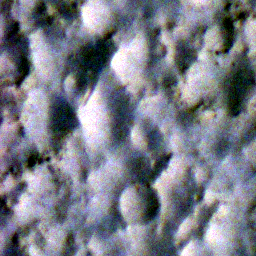}};
        \begin{scope}[x={(image.south east)}, y={(image.north west)}]
            \draw[white, line width=2pt] (0.05,0.06) -- (0.2075,0.06);
            \node[white, anchor=south] at (0.12875,0.075) {\small 10 m};
        \end{scope}
    \end{tikzpicture}
    \hfill
    \begin{tikzpicture}[baseline=(image.center)]
        \node[anchor=south west, inner sep=0] (image) at (0,0)
            {\includegraphics[height=3.75cm]{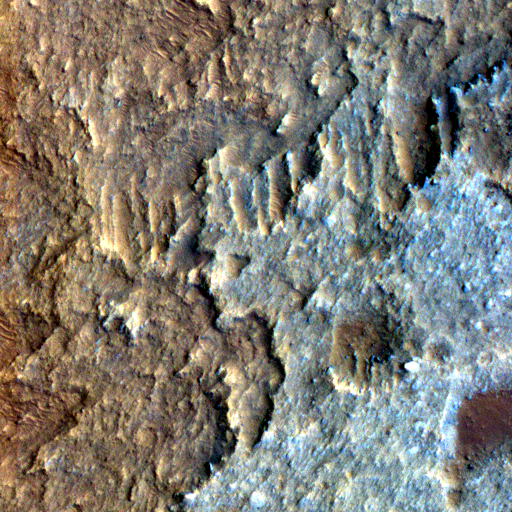}};
        \begin{scope}[x={(image.south east)}, y={(image.north west)}]
            \draw[white, line width=2pt] (0.05,0.06) -- (0.2009,0.06);
            \node[white, anchor=south] at (0.12545,0.075) {\small 20 m};
        \end{scope}
    \end{tikzpicture}

    \vspace{2ex}

    \begin{tikzpicture}[baseline=(image.center)]
        \node[anchor=south west, inner sep=0] (image) at (0,0)
            {\includegraphics[height=3.75cm]{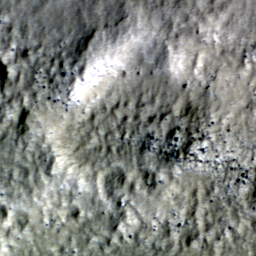}};
        \begin{scope}[x={(image.south east)}, y={(image.north west)}]
            \draw[white, line width=2pt] (0.05,0.06) -- (0.2069,0.06);
            \node[white, anchor=south] at (0.12845,0.075) {\small 20 m};
        \end{scope}
    \end{tikzpicture}
    \hfill
    \begin{tikzpicture}[baseline=(image.center)]
        \node[anchor=south west, inner sep=0] (image) at (0,0)
            {\includegraphics[height=3.75cm]{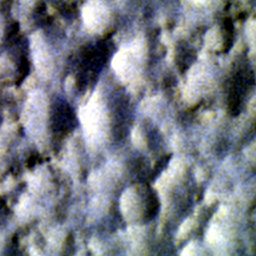}};
        \begin{scope}[x={(image.south east)}, y={(image.north west)}]
            \draw[white, line width=2pt] (0.05,0.06) -- (0.2075,0.06);
            \node[white, anchor=south] at (0.12875,0.075) {\small 10 m};
        \end{scope}
    \end{tikzpicture}
    \hfill
    \begin{tikzpicture}[baseline=(image.center)]
        \node[anchor=south west, inner sep=0] (image) at (0,0)
            {\includegraphics[height=3.75cm]{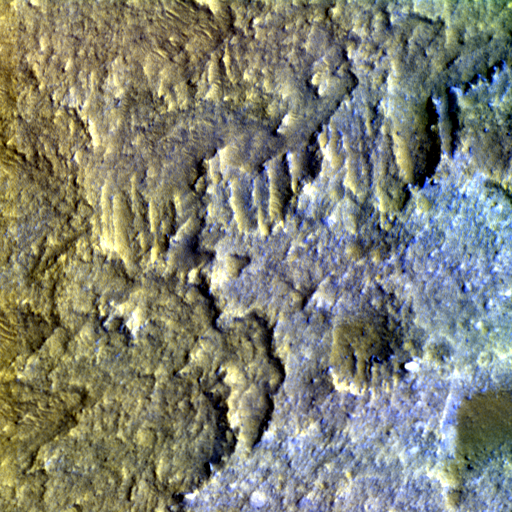}};
        \begin{scope}[x={(image.south east)}, y={(image.north west)}]
            \draw[white, line width=2pt] (0.05,0.06) -- (0.2009,0.06);
            \node[white, anchor=south] at (0.12545,0.075) {\small 20 m};
        \end{scope}
    \end{tikzpicture}

    \caption{Top row: True HiRISE images using the original RED4. Bottom row: HiPredict predictions for the same areas. Images are plotted in `IRB' format, meaning the infrared/red/blue-green bands are mapped to the red/green/blue channels to create the colour image, which helps to highlight subtle surface composition differences. Left column: ESP\_055232\_1595 ($-20.1668^\circ, 226.817^\circ\text{E}$). Middle column: ESP\_056705\_1440 ($-35.4419^\circ, 342.328^\circ\text{E}$). Right column: ESP\_012940\_1655 ($-14.0974^\circ, 318.282^\circ\text{E}$).}
    \label{fig:Examples}
\end{figure}

To capture the diversity of priorities for the Mars science community, we also present two other metrics. All metrics are defined in Section \ref{sec:val_metrics}. Structural Similarity Index (SSIM) is useful for assessing how similar the high-frequency spatial variation of a predicted image is to the ground truth. HiPredict (0.725) outperforms SYN4 by 9\% on this metric. For applications where accurate pixel measurement is important, a more relevant metric is the proportion of pixels that are within 1\% of their true value (PWT 1\%). PWT is approximately two times higher for HiPredict (20.844\%) compared with SYN4. The full metrics, as well as those for previous model versions described in Section \ref{sec:other_models} are seen in Table \ref{tab:metrics} with example images in Figure \ref{fig:Examples}. As SYN4 is prone to outliers and a wide distribution of results, we present median values alongside the mean to show where large outliers may skew mean results. 

\begin{table}[htbp]
\centering
\caption{Performance metrics for unseen data. Arrows indicate if higher or lower values are better.}
\label{tab:metrics_evaluation}
\begin{tabular}{lcccccc}
\toprule
& \multicolumn{2}{c}{\textbf{NMAE (\%) $\downarrow$}} & \multicolumn{2}{c}{\textbf{SSIM $\uparrow$}} & \multicolumn{2}{c}{\textbf{PWT 1\% $\uparrow$}} \\
\cmidrule(lr){2-3} \cmidrule(lr){4-5} \cmidrule(lr){6-7}
\textbf{Name} & \textbf{mean} & \textbf{median} & \textbf{mean} & \textbf{median} & \textbf{mean} & \textbf{median} \\
\midrule
Linear & 4.012 & 3.508 & 0.542 & 0.596 & 4.926 & 2.632 \\
Linear with meta & 4.234 & 3.594 & 0.544 & 0.588 & 4.851 & 2.634 \\
Basic UNet & 2.064 & 1.339 & 0.687 & 0.747 & 13.841 & 12.750 \\
Customised UNet & 6.453 & 2.990 & 0.632 & 0.690 & 6.729 & 2.924 \\
SYN4 & 3.118 & 1.545 & 0.663 & 0.722 & 11.069 & 9.148 \\
\textbf{HiPredict} & \textbf{1.110} & \textbf{0.943} & \textbf{0.725} & \textbf{0.788} & \textbf{20.844} & \textbf{18.801} \\
\bottomrule
\label{tab:metrics}
\end{tabular}
\end{table}

We also examined potential areas where the SYN4 models may be weak by examining underlying assumptions. We have quantified this by measuring the standard deviation of SYN4 predictions and we see the standard deviation for the NMAE decrease from 6.50 with SYN4 to 0.93 with HiPredict. We see modest improvements from SSIM and PWT 1\% metrics, showing the HiPredict model has a lower variance and fewer large outliers relative to SYN4. This is illustrated in Supplemental Information S1.

We also plot the remaining loss as `diff' plots for each figure. We observe that the plot of $|True - HiPredict|$ consistently contains little high-frequency detail and is most commonly a `salt-and-pepper' type image. This suggests the remaining error is largely instrument noise or binning artefacts, indicating either a floor to the predictability of these images, or the need for more advanced metadata modelling and noise handling. Diff plots from all images are included in Supplemental Information S4.

\subsection{Analysis of colourful/complex images}

\begin{figure}[h]
    \centering
    \includegraphics[width=0.7\textwidth]{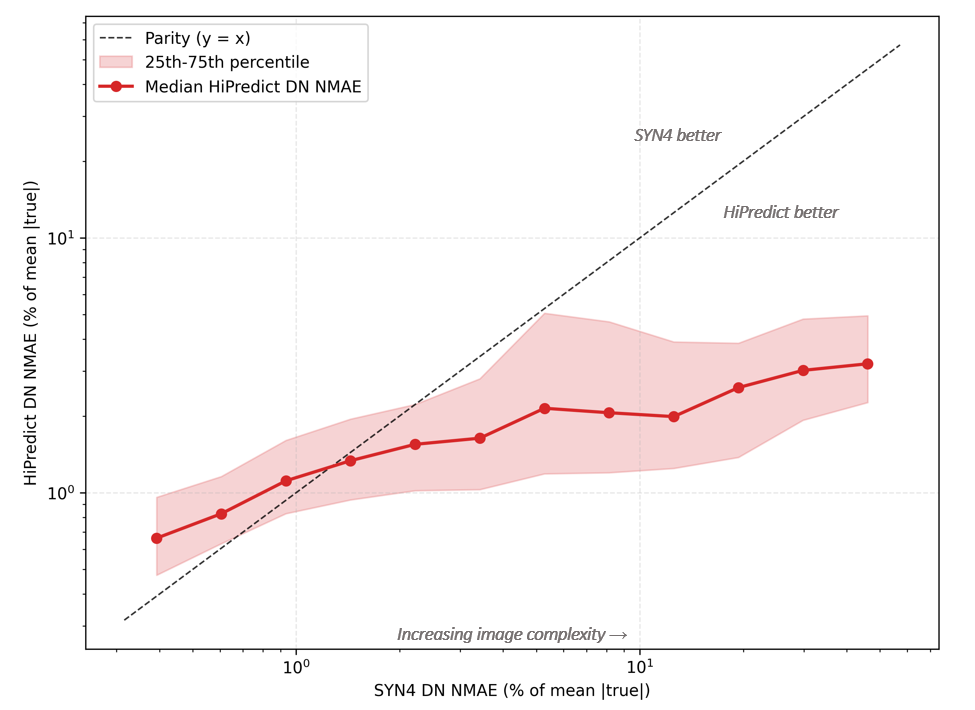}
    \caption{Using high SYN4 error as a proxy for colourful and complex images, we can evaluate HiPredict performance on this subset. }
    \label{fig:complexity_w_err}
\end{figure}

As SYN4 is the IR10 band linearly scaled to match RED4, we can consider patches with a high SYN4 error to be more colourful and complex than the average observation. As such, evaluating HiPredict on observations with a high SYN4 error acts as a measurement of the model's performance on a more challenging subset of images. Figure \ref{fig:complexity_w_err} illustrates that as image complexity increases, HiPredict error does not increase at the same rate, remaining lower. For images where the SYN4 NMAE is over 10\%, HiPredict exhibits an average 7.7 times improvement between the two models for this metric.

\subsection{Robustness across different operating and environmental conditions}

\begin{wrapfigure}{L}{0.6\textwidth}
    \centering
    \includegraphics[width=\linewidth]{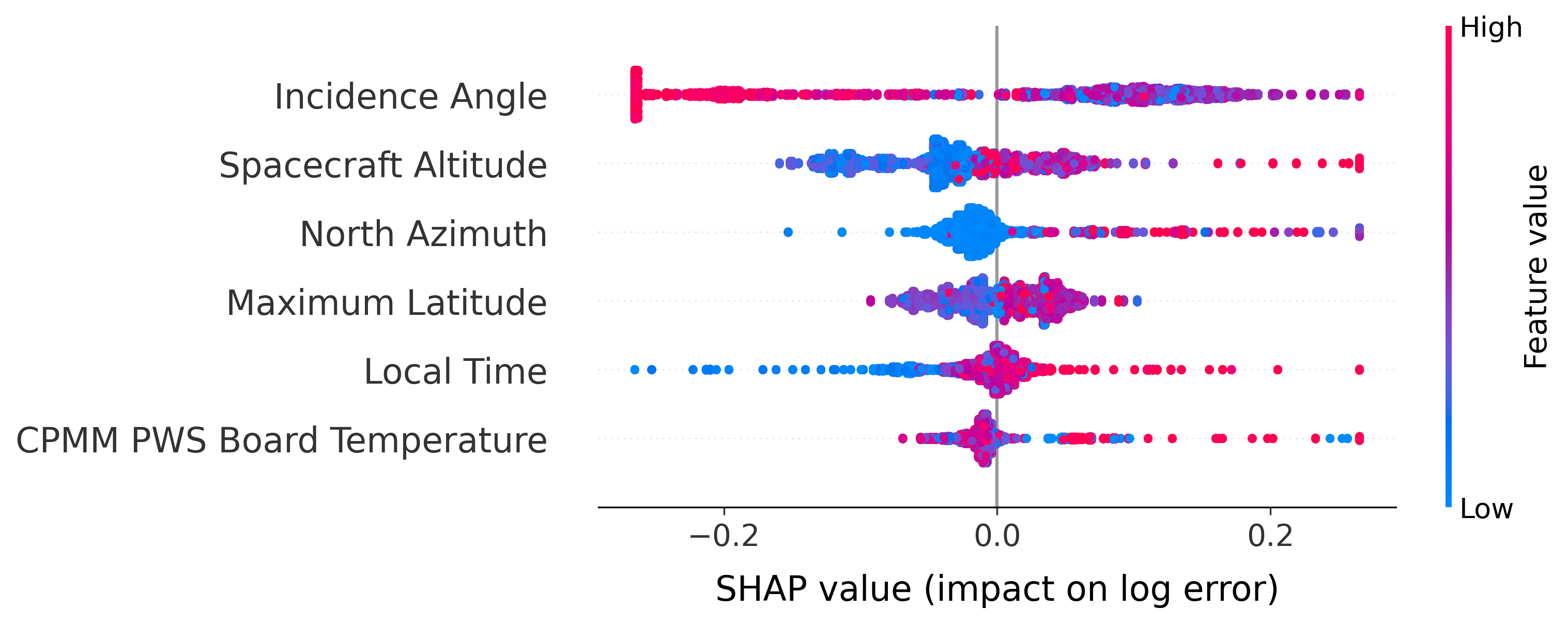}
    \caption{Log-transformed SHAP analysis of six most consequential metadata values provided to the model, excluding the top 10 outliers. Points to the left of center have a positive impact on model performance, points to the right a negative impact. Red points indicate a larger value for that metadata item and blue a lower value. Verticality indicates density of data points. The complete SHAP plot can be seen in Supplemental Information S6}
    \label{fig:shap}
\end{wrapfigure}

While metadata values are included in the model for training and inference, they can still affect model performance. We assess the impact of specific metadata values on model performance across 474 held-out observations. 

To assess the impact of metadata, SHAP analysis was conducted. Across these observations, the HiPredict loss is heavily skewed (skew of 5.2) with the worst 10 observations carrying 74\% of the squared deviation. On these observations, SYN4 has a mean loss of 0.33 against an overall mean of 0.076 showing the images are intrinsically difficult for both models. Therefore, the analysis was conducted three ways: with the worst 10 excluded, with a log-transformed loss and with both.

Using this method, we find only three features hold a top-6 position across all three SHAP variants: incidence angle, spacecraft altitude and north azimuth. Under the strictest test, incidence angle (the solar zenith angle at acquisition) dominates at 0.144 in log-error units, which is 2.6 times the next feature. This is likely a proxy for scene contrast and dynamic range as high incidence angles are usually higher contrast with lower colour variation due to more shadows and shading differences. 

The other two are spacecraft altitude and north azimuth. Spacecraft altitude likely relates to the spatial scale of surface features. Higher altitude corresponds to higher error likely as coarser pixels integrate more surface material and inter-band relationships become more complex. North azimuth can be seen as a proxy for latitude in near-polar orbits and so we see from this analysis that polar images are often more poorly predicted. 

The log-transformed SHAP plot which excludes the top 10 outliers can be seen in Figure \ref{fig:shap}. It shows the top six rows but only the top three are robust, positions 4-6 (maximum latitude, local time and CPMM board temperature) vary with the other two SHAP variants. 

Furthermore, the model was trained and fully converged with training dataset sizes of 20,000, 50,000 and the maximum 81,717 samples. The median NMAE values of these runs on the test set are 1.46\%, 1.02\% and 0.93\% respectively showing that increased dataset size does improve model performance. This is confirmed by a similar study with the linear model, showing increased performance with increased data volume, although no linear models outperformed SYN4 (Table~\ref{tab:metrics}). We also see diminishing returns between the two larger datasets suggesting that the dataset size is sufficient. Interestingly, comparing the runs with the largest and smallest dataset, the SSIM only improves by 2\% (0.7109 to 0.7254) whereas the brightness prediction improves significantly with the larger model. The average brightness offset increases nearly sixfold from 0.217\% in the best performing model to 1.259\% in the worst. As the IR10 and BG12 inputs contain much of the spatial information present in RED4, this likely needs much fewer samples whereas the radiometric information is likely harder to model and likely requires more samples.

\subsubsection{Model error decomposition}

Residuals were decomposed into mean offset (DC), low-frequency structure (LF), and high-frequency detail (HF) to identify what limits model performance. The median image is dominated by HF error (69.6\%; IQR 47--83\%), while LF and DC terms are small. Thus, for most scenes, the model recovers the broad radiometric and structural content, and the remaining error is mainly fine-scale detail that is weakly constrained by the available colour-band context. We believe much of this is due to unrecoverable underlying instrument noise, or the IR10 channel being binned higher than the RED4 label it is trying to predict.

The mean error shows the opposite pattern, with LF contributing 66.7\% of total residual energy. This discrepancy indicates that the validation loss is dominated by a small number of difficult images rather than the typical case. These failures are primarily smooth radiometric or structural errors, not high-frequency noise. The DC term is small in aggregate (6.4\%) but variable between images (IQR 2--29\%) so individual brightness offsets remain, but do not produce a significant systematic dataset-wide bias. This trend of a small number of outliers skewing the mean loss is why both median and mean values are reported in Table \ref{tab:metrics}.

\subsubsection{Improvements in spatial structure}

Although a portion of the high-frequency residual is irreducible (see Supplemental Information S4), the model still recovers structure that the SYN4 baseline cannot, which accounts for its higher SSIM in Section \ref{sec:val_accuracy_and_baseline_comparison}. SYN4 applies a single global multiplicative scale to the IR10 band, preserving its spatial structure exactly; it therefore has no flexibility to alter local contrast or sub-pixel detail. Where the colour bands are binned more coarsely than RED, each native IR10 pixel is upsampled to a flat block on the RED grid that contains no genuine high-frequency content, and the global scaling leaves these blocks unmodified. A convolutional model, such as HiPredict, applies a spatially-varying, per-pixel transformation and additionally draws on the full-resolution neighbouring RED CCDs, allowing it to infer plausible fine-scale structure correlated with the target band.

\subsection{Conversion to full strips}

In addition to bulk analysis on the patches, 40 observations were assessed as full HiRISE strips. As the model is fully convolutional, no modifications are required between inference on 256x256 training patches and real strips tens of thousands of pixels long. 

For the evaluation images, the NMAE for the strip actually improved relative to the patches, decreasing from 1.27\% to 0.96\%. The SYN4 loss also improved when applied to the full strip rather than a patch, decreasing from 2.65\% for this subset to 1.66\%. This decrease is expected to be due to the positive effect of the neighbour kernels in images much larger than their height.

%% file: Sections/5.Downstream.tex
\section{Downstream task evaluation}
\label{sec:results_downstream}

As discussed in Section \ref{method:downstream_eval}, we split our evaluation into three tiers. Section \ref{sec:results} discusses various aspects of model performance across the broad domain of the full test set. In this Section, we evaluate the model against criteria specific to Mars science.

First, we assessed our model against several subsets of images, each corresponding to a science theme, to understand its performance in regions of significance. A total of 10 examples of icy craters were assessed, with HiPredict reporting an average NMAE of 1.68\% compared to 9.95\% for SYN4. For polygonal ridge spectra, we similarly see a significant drop from 5.16\% to 1.32\% with our model across 8 examples. Chloride deposits were more competitive between the two models with the SYN4 average across 8 examples reported as 3.03\% compared to 1.37\% for HiPredict. 

The order of this performance follows the order of local spectra discontinuity which is strongest in icy impacts, substantial in polygonal ridges and smoother in chloride deposits. Because SYN4 predicts the RED band from IR using simple scaling, it fails increasingly as the relationship between these bands diverges from the background. Our model is less vulnerable as it uses spatial context and same-band information rather than relying on spectral similarity. The full list of observations and coordinates used are available in the Supplemental Information S5.

For further analysis, we focused on icy impact craters. Pixels covering icy material were selected and adjusted for background noise following \citeA{RANGARAJAN2024115849} and these pixel values were then compared to background pixels, which represent dry, dusty, non-ice bearing material, using the same correction. The spectra between HiPredict, SYN4 and the ground truth (labelled as RED4) are then compared in Figure \ref{fig:icy-examples}.

We find SYN4 consistently under-estimates the pixel values, causing a peak in BG/RED and WATER-ICE band ratio parameters. This is consistent across all images and visualised by the blue lines in Figure \ref{fig:subfig:icy-spectra}. The under-estimation that we see is illustrated by a reduction of both the mode of the distribution and values typically at the higher end of the histogram, which typically correspond to any icy pixels present and suggests that IR is a worse indicator at high RED DN values than low values as we see these pixels being more poorly predicted. This effect is not as pronounced at the lower ends of the histogram; the spectra/parameters for ice-poor background pixels (right column in Figure \ref{fig:subfig:icy-spectra}) look very consistent across the three data products. The spectra and band ratios produced with HiPredict achieve a nearly identical performance in identifying water ice-bearing pixels as the original RED4 images, while the spectra and band ratios produced with SYN4 do not. This suggests HiPredict is capable of reliable identification of water-ice, a task which SYN4 is currently unable to do.

\newcommand{\hirisewithscale}[4]{%
    \begin{tikzpicture}
        \node[anchor=south west, inner sep=0] (image) at (0,0)
            {\includegraphics[width=0.29\textwidth]{#1}};
        \begin{scope}[x={(image.south east)}, y={(image.north west)}]
            \draw[white, line width=2pt] (#2,0.06) -- (#3,0.06);
            \node[white, anchor=south] at (#4,0.075) {\small 10 m};
        \end{scope}
    \end{tikzpicture}%
}

\newcommand{\spectraplot}[2]{%
    \begin{tikzpicture}
        \node[anchor=south west, inner sep=0] (image) at (0,0)
            {\includegraphics[width=0.29\textwidth]{#1}};
        \begin{scope}[x={(image.south east)}, y={(image.north west)}]

            \fill[white] (#2,0.8) rectangle (1.00,1.00);

            \node[
                anchor=north east,
                fill=white,
                fill opacity=0.95,
                text opacity=1,
                inner sep=0.02pt,
                font=\normalfont\rmfamily\fontsize{5}{5.5}\selectfont,
                align=left
            ] at (1.2,0.96) {%
                \textcolor{blue}{\rule[0.4ex]{0.9em}{0.35pt}}~SYN4\\
                \textcolor{red}{\rule[0.4ex]{0.9em}{0.35pt}}~RED4\\
                \textcolor{black}{\rule[0.4ex]{0.9em}{0.35pt}}~HiPredict
            };

        \end{scope}
    \end{tikzpicture}%
}

\begin{figure}[htbp]
    \centering

    \subcaptionbox{%
        Left: Selected pixels marked on real HiRISE images, with placement following \citeA{RANGARAJAN2024115849}; the bright, bluish material is exposed water ice. Middle: Model predictions for the icy pixel. Right: Model predictions for the background pixel. The top observation is taken from observation ESP\_051070\_1285 ($51.2757^\circ, 101.053^\circ\text{E}$) and the bottom from ESP\_045973\_2440 ($63.9407^\circ, 282.389^\circ\text{E}$).%
        \label{fig:subfig:icy-spectra}
    }{%
        \begin{minipage}{\textwidth}
            \centering

            \hirisewithscale{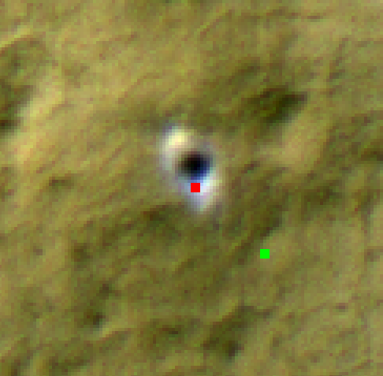}{0.05}{0.20625}{0.128125}\hfill
            \spectraplot{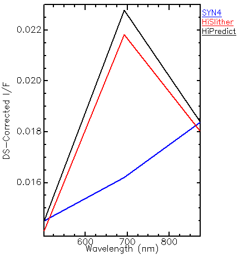}{0.845}\hfill
            \spectraplot{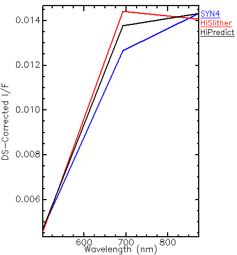}{0.84}

            \vspace{1ex}

            \hirisewithscale{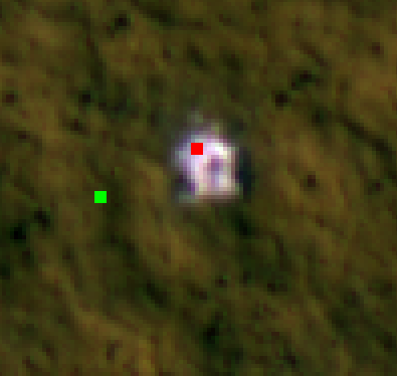}{0.05}{0.1760}{0.1130}\hfill
            \spectraplot{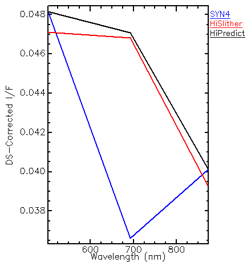}{0.835}\hfill
            \spectraplot{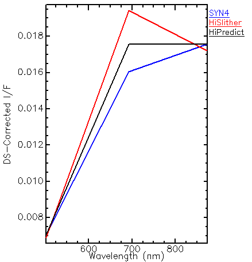}{0.825}
        \end{minipage}
    }

    \vspace{0.5ex}

    \subcaptionbox{%
        A BG/RED versus WATER-ICE spectral parameter plot for the icy-impact examples. HiPredict (black square) aligns much more closely to the true values (RED4, red upward triangle, following \citeA{RANGARAJAN2024115849}) than SYN4 (green downward triangle). Points in the top-right quadrant are considered reliable water-ice detections (“ice-rich”).%
        \label{fig:icy-plot}
    }{%
        \includegraphics[width=0.8\textwidth]{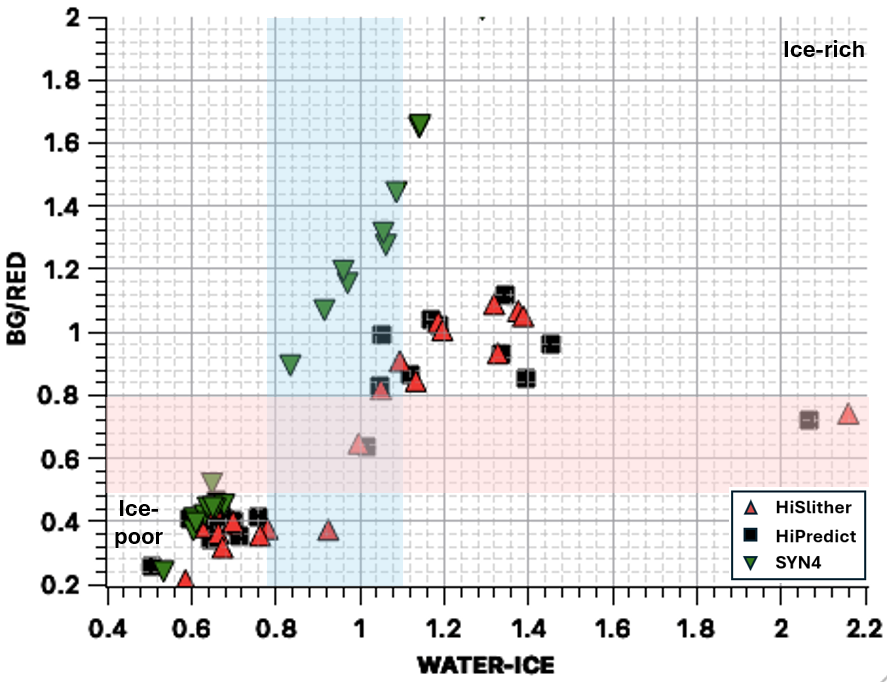}
    }
    
    \caption{A deeper comparison of the models using icy impacts as examples.}
    \label{fig:icy-examples}
\end{figure}

%% file: Sections/6.Discussion.tex
\section{Discussion}
\label{sec:discussion}

In this work, we have shown that spectral information between different bandpasses is learnable. In the future, these relationships between spectral bands could be extended to detectors without ground truth data for colour predictions across wider arrays (such as RED3 and RED6 predictions). For example, HiRISE-style focal planes could be designed in the future with the capability of using similar ML-based models to fill in the full swath with predicted 3-band colour in mind. This would require only narrow ground-truth data and hence provide space, power and compute savings. Variations of this model could also be used for load mitigation, for example some observations could be imaged with one fewer band, with the missing band predicted downstream, reducing the load on the instrument. If combined with data from similar sources (such as the Colour and Surface Stereo Imaging System -- CaSSIS), or lower resolution ones (such as the Mars Colour Imager -- MARCI), this work could lay the foundation for spectral interpolation across resolutions and instruments, creating `virtual satellites' for Mars similar to many Earth-based proposals \cite{brownAlphaEarthFoundationsEmbedding}.

This method does, however, show clear constraints. Firstly, the volume of planetary data available creates limitations from a compute perspective. Less than 20\% of HiRISE images were used for training, likely covering less than 1\% of the surface of Mars, but this dataset size was over 200 GB. The size and shape of HiRISE images also imposed constraints on the design choices for our model. Full-size HiRISE images are of variable sizes and are extremely large so non-convolutional methods (such as diffusion or other generative models) would have been difficult to implement and would have limited normalisation methods within hidden layers of the model. 

Future work will expand the available training data, and experiment with other forms of Convolutional Neural Networks such as pre-encoding certain inputs or making the model bi-directional to further enable the model to capture underlying relationships. Also, more detailed probes could be applied to this model (such as principal component analysis) to better define the current learned relationship between different inputs and RED bandpass. 

\section{Conclusions}
\label{sec:conclusion}

Here, we introduce the HiPredict model: a machine learning model able to predict missing imagery from the RED4 detector of the HiRISE camera, following its partial failure in 2023. Our overall error is nearly three times smaller than the baseline model, SYN4 (a brightness-adjusted version of the overlapping IR10 image), with both spatial and radiometric improvements. 

We validate the model first across a wide test set, then use standard evaluation metrics on a subset of images relating to specific science themes and finally evaluate deeply on one example case, recreating a full scientific pipeline. We find HiPredict remains more accurate than SYN4 and captures spectral variations missed by the baseline product. Our deepest evaluation case, predicting icy impact craters, demonstrates HiPredict's ability to identify water-ice reliably -- a task SYN4 is not able to do. 

The development of HiPredict has several direct benefits. Firstly, it is able to restore the missing RED4 strip with updated predictions. Not only does this allow the infilling of single-band RED products (to a higher accuracy than SYN4) but it also creates the opportunity for three-band colour images in this swath, a task SYN4 is unable to fulfil as it is directly derived from another band. We supplement these predicted products with detailed performance metrics in different areas. 

In addition to the practical benefits, this work shows that spectral interpolation can be achieved with a competitive accuracy which has strong implications for instrument design and planetary imaging in the future, especially around supporting remote-sensing missions as they age.